\documentclass[10pt,journal,compsoc]{IEEEtran}
\usepackage{amsmath,amsfonts}
\usepackage{array}
\usepackage{textcomp}
\usepackage{stfloats}
\usepackage{url}
\usepackage{verbatim}
\usepackage{graphicx}
\usepackage{cite}
\usepackage{amsthm}
\usepackage{subcaption}
\usepackage{hyperref}
\usepackage[numbers]{natbib}  
\hypersetup{colorlinks=true, linkcolor=blue, citecolor=blue, urlcolor=blue}  

\usepackage{booktabs}
\usepackage{multirow}
\usepackage{makecell}
\usepackage{bm}
\usepackage{mathrsfs}
\usepackage{amssymb} 
\usepackage[inkscapelatex=false]{svg}

\usepackage[table]{xcolor}
\usepackage{color}

\newcommand{\rgba}[1] {\small({\color[rgb]{0.3,0.5,0.8}#1})}
\newcommand{\rgbb}[1] {({\color[rgb]{0.3,0.0,0.1}#1})}

\usepackage[ruled,vlined,linesnumbered]{algorithm2e}

\usepackage[switch]{lineno}  

\newif\iflinenumbers
 \linenumbersfalse 

\iflinenumbers
\linenumbers
\modulolinenumbers[1] 
\fi

\newtheorem{theorem}{Theorem}[section]  

\begin{document}

\title{Personalized Federated Learning via Gaussian Generative Modeling}

\author{
	Peng Hu, Jianwei Ma
	\thanks{
		P. Hu is with the School of Mathematics, Harbin Institute of Technology, Harbin, China.
	}
	\thanks{
		J. Ma is with the Institute for Artificial Intelligence and the School of Earth and Space Sciences, Peking University, Beijing, China, and also with the Institute for Artificial Intelligence and the School of Mathematics, Harbin Institute of Technology, Harbin, China.
	}
	\thanks{
		Corresponding author: J. Ma (e-mail: jwm@pku.edu.cn).
	}
}

 \IEEEtitleabstractindextext{
\begin{abstract}
Federated learning has emerged as a paradigm to train models collaboratively on inherently distributed client data while safeguarding privacy. In this context, personalized federated learning tackles the challenge of data heterogeneity by equipping each client with a dedicated model. A prevalent strategy decouples the model into a shared feature extractor and a personalized classifier head, where the latter actively guides the representation learning. However, previous works have focused on classifier head-guided personalization, neglecting the potential personalized characteristics in the representation distribution. Building on this insight, we propose pFedGM, a method based on Gaussian generative modeling. The approach begins by training a Gaussian generator that models client heterogeneity via weighted re-sampling. A balance between global collaboration and personalization is then struck by employing a dual objective: a shared objective that maximizes inter-class distance across clients, and a local objective that minimizes intra-class distance within them. To achieve this, we decouple the conventional Gaussian classifier into a navigator for global optimization, and a statistic extractor for capturing distributional statistics. Inspired by the Kalman gain, the algorithm then employs a dual-scale fusion framework at global and local levels to equip each client with a personalized classifier head. In this framework, we model the global representation distribution as a prior and the client-specific data as the likelihood, enabling Bayesian inference for class probability estimation. The evaluation covers a comprehensive range of scenarios: heterogeneity in class counts, environmental corruption, and multiple benchmark datasets and configurations. pFedGM achieves superior or competitive performance compared to state-of-the-art methods.



\end{abstract}


}

\maketitle

\section{Introduction}
\IEEEPARstart{D}{ue} to constraints such as privacy protection and data security, traditional data aggregation is often prohibitive in many scenarios. In such cases, multiple clients hold small local datasets that remain isolated in silos, preventing any data expansion to enhance the performance of Deep Neural Networks (DNNs) \cite{Tan2023silos, Liu2021Privacy, Pfeiffer2023Survey, Kaissis2021privacy, Du2020Issues}. Through enabling privacy-preserving collaborative training across distributed datasets, Federated Learning (FL) effectively unlocks the latent value of decentralized data, providing a robust solution to the challenge of data silos in deep learning applications\cite{Yang2018Applied, Sheller2019Brain, Ramaswamy2019Emoji, Majcherczyk2021Flow, Wang2022ATPFL, Mistry2024Emoji, Chen2025FedMEM}. For example, the pioneering FedAvg\cite{McMahan2017FedAvg} avoids the transmission of raw data by aggregating client models over multiple rounds. With clients updating local models and a central server maintaining a global one, collaboration is facilitated by the infrequent exchange of model parameters. Such a strategy has been proven effective when the data across clients is independent and identically distributed (IID) \cite{Stich2019FedSGD}. However, in the presence of client-side data drift, despite substantial efforts\cite{Li2020Optimization, Li2020Convergence, Khaled2020Convergence, Karimireddy2020Stochastic, Wang2020Optimization}, a single global model often struggles to generalize effectively across non-IID data distributions.

Personalized Federated Learning (PFL) serves as a novel approach for tackling this challenge. It strikes a balance by building a personalized model for each client: the local model contributes to the global aggregation while simultaneously benefiting from collaborative training. The customized model addresses data heterogeneity caused by varying client-side data collection preferences\cite{Othmane2021fedMD, Xia2025pFedMLKD}, such as feature distribution shift, label distribution shift, and data imbalance. However, excessive personalization can hinder knowledge fusion across clients, thereby undermining the federated mechanism. A range of techniques, such as regularizing local objectives\cite{Dinh2020pFedMe, Li2021Ditto}, meta-learning\cite{Jiang2019Meta, Fallah2020Meta}, local-global parameter interpolation \cite{Deng2020APFL} and decoupling representation from classifier learning\cite{Chen2022FedRoD, Li2024pFedFAFT}, have been developed with the aim of promoting global aggregation while maintaining high performance on local data. Substantial research focuses on data heterogeneity arising from class imbalance, while feature distribution shift (e.g., differences in noise levels) has received only limited exploration.


\begin{figure}[htbp]
	\centering
	\includegraphics[width=0.45\textwidth]{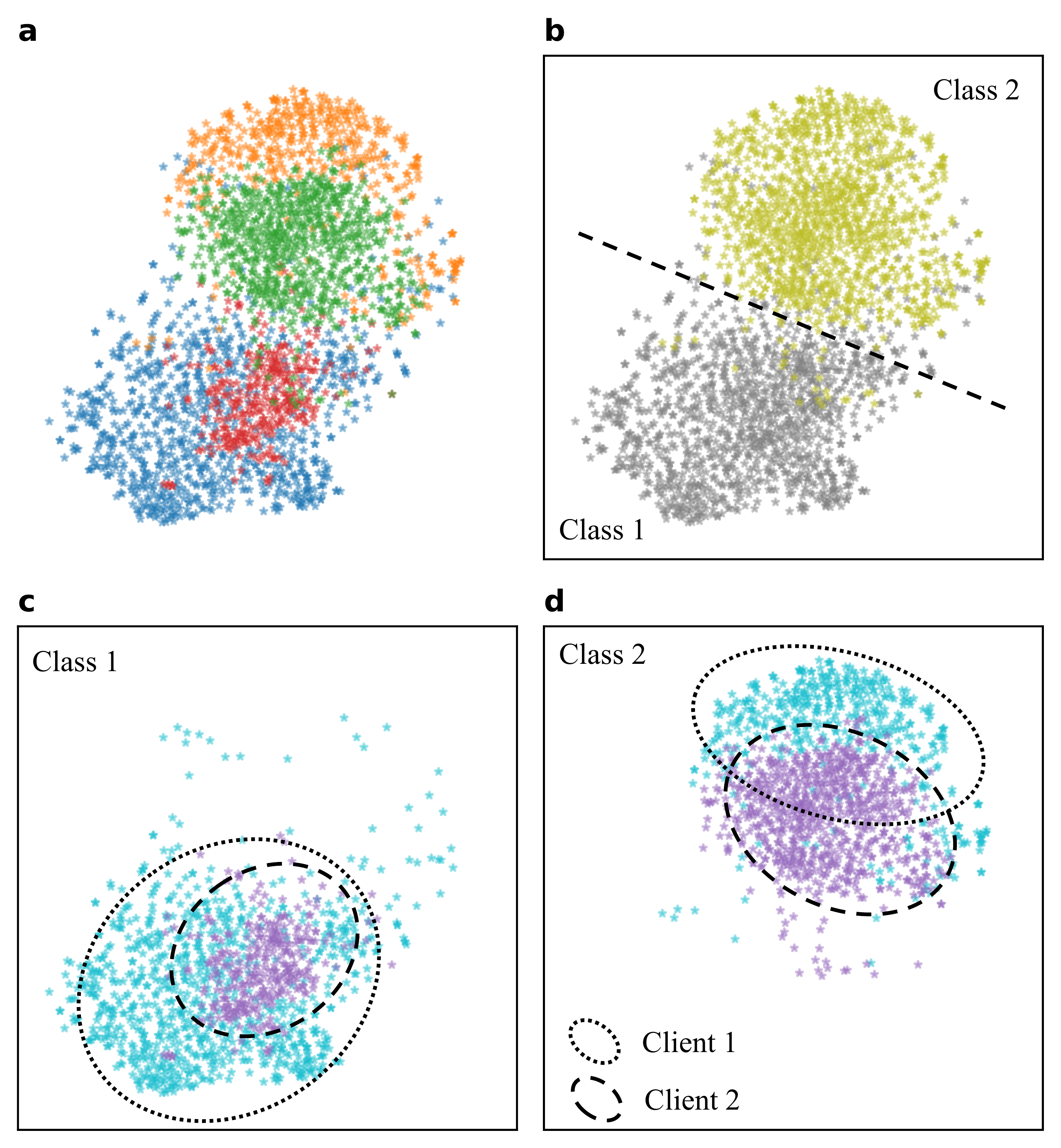}  
	\caption{ t-SNE visualization of representation distribution divergence caused by client data heterogeneity. (a) Representations of two classes from two clients; (b) Colored by class; (c, d) Colored by client.  Subfigure (c) displays the feature representations of class 1, while (d) displays those of class 2. Different clients exhibit distinct cluster means and covariance structures. }
	\label{fig_tsne_look}
\end{figure}

In this work, we focus on the classification task. It can be viewed as a form of nonlinear compression that maps from a high-dimensional pixel space to a low-dimensional class space. Under the scenario of feature shift, directly training a network using only class labels appears overly simplistic, as the over-compressed resulting representations cannot reflect underlying data heterogeneity. Motivated by the success of decoupled representation-classifier learning\cite{Xu2023FedPAC, McLaughlin2024pFedFDA}, we reformulate the classification task within a representation-based generative modeling framework. We expect that the intricate local data heterogeneity, which is difficult to model directly, is well-reflected within the representation space. Then, each client can construct an effective personalized classifier based on its well-expressed representation distribution. Moreover, information-rich representations of client data empower the formulation of a wider array of personalized learning objectives.

Fig. \ref{fig_tsne_look} shows the representation distribution when simulating client data heterogeneity by applying motion blur to Client 1 and fog noise to Client 2. It can be observed that different clients within the same class exhibit distinct distribution characteristics, including different cluster means and covariance structures. This inspires us to adopt a generative modeling approach, constructing training objectives based on the representation distributions.


To achieve easily separable data representations, we employ a Gaussian-based framework. Assuming that the neural network maps images of the same class to a Gaussian, the representation distribution for multiple classes then forms a Gaussian mixture. In this case, the class of an image can be inferred by identifying its associated Gaussian component in the representation space, based on class-conditional probability. To simulate statistical heterogeneity across clients, the image data for each class on a client is assumed to be generated via re-sampling from an original distribution. We further assume that this re-sampling weight is also proportional to the Gaussian. Consequently, the distribution in the representation space for each client is also Gaussian, with client heterogeneity captured by distinct mean and covariance parameters

\begin{figure*}[htbp]
	\centering
	\includegraphics[width=1\textwidth]{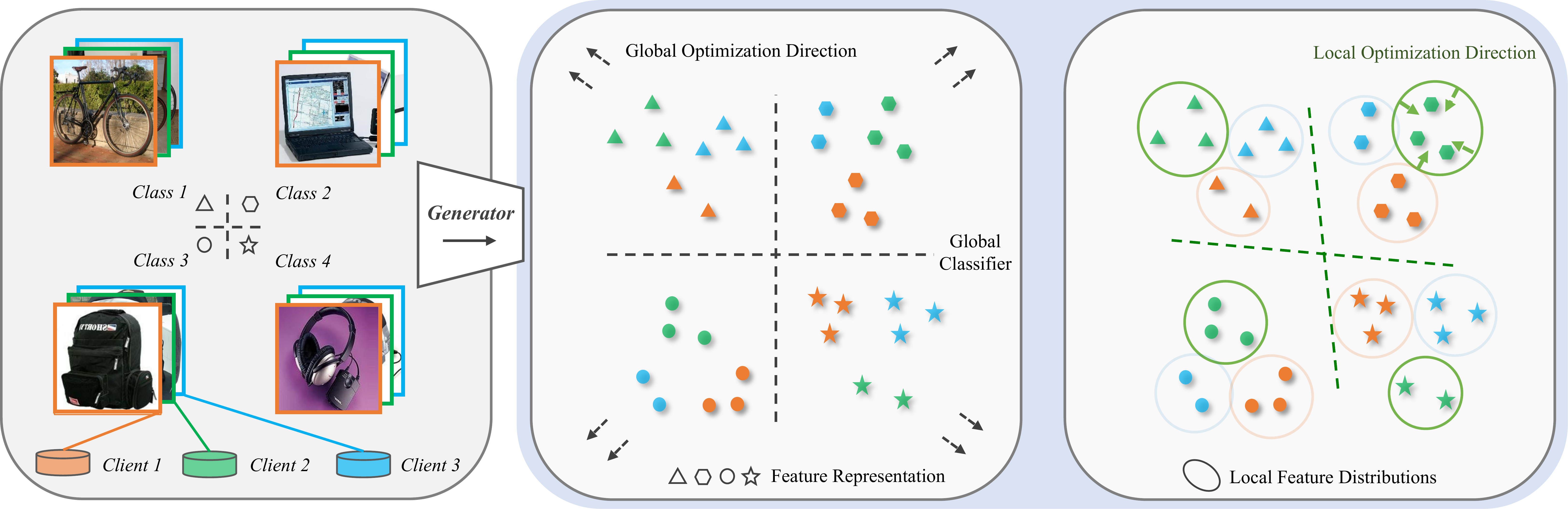}  
	\caption{Overview of the generator training. (Left) Data from different clients of the same class exhibit heterogeneity, and consequently, their distributions diverge in the representation space. (Middle) The shared objective drives features of different classes to diverge along distinct directions, whereas (right) the client personalized objective prompts features of the same class to aggregate around client-specific centers. }
	\label{fig_pFedGM}
\end{figure*}

An intuitive way for more accurate classification involves two complementary goals: maximizing the distance between class means and minimizing the variances within each class. To achieve the collaboration-personalization trade-off, we introduce a global-local collaborative method which push the global inter-class means away from each other (Fig. \ref{fig_pFedGM} middle) and shrink the client intra-class variances (Fig. \ref{fig_pFedGM} right). The shared objective originates from the server, encouraging the models to learn common features, while the client-customized distributions and objectives foster adaptation to local data. These two components jointly promote the updating of the model parameters.


Guided by the Gaussian mixture assumption, our approach employs a two-level model decoupling mechanism to attain the training objective. The initial step is to decouple the model into a generative model and a Gaussian classifier, following the earlier description. Subsequently, the Gaussian classifier head is further decoupled into a statistics extractor and a navigator, where the navigator adaptively determines the global optimization direction. This allows for compatibility with downstream client adaptation without introducing extra parameters.


Following the training of the generative model, we apply a dual-scale fusion method to construct personalized classifiers. The global and local feature parameters are coupled via an adaptive mechanism, motivated by the Kalman gain \cite{Simon2006Kalman}, enabling the global estimate to adapt to local representation distributions while mitigating overfitting with limited local data.

\textbf{Contributions:} We propose a novel PFL framework based on Gaussian generative modeling (pFedGM). To simulate data heterogeneity, we construct a client-personalized representation distribution. Generative modeling is used to derive shared and local objectives that balance collaborative training and client personalization. Our contributions can be summarized as follows:

\begin{itemize}
	\item We introduce a new perspective for PFL by modeling data heterogeneity through a client-level representation distribution and a Gaussian re-sampling strategy.
	\item A new PFL approach is proposed, using inter-class and intra-class representations to balance collaborative training with client personalization. And a dual-scale fusion method based on information gain is introduced for personalized classifier adaptation.
	\item Extensive experiments on natural image classification under various heterogeneity settings demonstrate the advantages and robustness of pFedGM.
\end{itemize}

\section{Related Work}
The performance of standard FL algorithms, such as FedAvg\cite{McMahan2017FedAvg}, is severely hampered by statistical heterogeneity (non-IID data) across clients. This heterogeneity induces client drift, thereby posing the well-documented challenge of slow convergence and degraded performance. In response to this issue, considerable research has focused on improving global model learning under non-IID data settings. To improve client-side local training, prior efforts have introduced techniques such as adding regularization to the local loss function \cite{Acar2021Regularization} or applying corrections to mitigate update bias\cite{Karimireddy2020SCAFFOLD, Murata2021Bias, Dandi2022Implicit}. Strategies such as class-balanced re-sampling and loss re-weighting are applied to mitigate the effects of imbalanced data distributions across clients and improve training \cite{Hsu2020Data, Wang2021Addressing}. Another line of research aims to accelerate convergence and mitigate the negative impact of data heterogeneity by selecting clients that contribute more significantly to the global model during the aggregation phase \cite{Wang2020Optimizing, Tang2022FedCor, Wu2022Node, Fraboni2021Clustered}. Additional strategies include data sharing and augmentation \cite{Zhao2018Federated, Yoon2021FedMix}, contrastive learning\cite{Li2021Contrastive}, knowledge distillation\cite{Lin2020FedDF, Zhu2021Data}, and prototype learning \cite{Michieli2021Prototype, Tan2021FedProto, Mu2023fedproc, Ye2023FedFM, Zhou2024FedFA}. Although these methods continuously improve performance on Non-IID data, a single global model struggles to achieve optimal performance across all clients with highly divergent data distributions.

PFL addresses this by shifting from ``model uniformity" to ``solution adaptability", learning a dedicated model for each client that is closely aligned with its local data distribution. Its central challenge lies in striking a balance between leveraging global data for collaborative training and achieving precise adaptation to local distributions. For instance, clustered federated learning groups similar clients together and learns multiple group-level global models\cite{Mansour2020Three, Duan2021Flexible, Sattler2021Clustered, Ghosh2022An}. Other popular approaches include meta-learning-based local adaptation\cite{Jiang2019Meta, Fallah2020Meta}, multi-task learning with model discrepancy penalties\cite{Smith2017FedMTL, Li2021Ditto}, client-specific model aggregation\cite{Zhang2021Order, Huang2021cross, Beaussart2021WAFFLE}, and the decoupling of feature extractors and classifiers\cite{Collins2021FedRep, Oh2022FedBABU, Xu2023FedPAC, McLaughlin2024pFedFDA}. 

Our work is most closely related to studies on decoupling of feature extractors and classifiers. The main distinction among these methods lies in how the client-specific classifier heads are acquired. This, in turn, influences the training of the representation learning process. For example, FedBABU\cite{Oh2022FedBABU} proposes to train the feature extractor using a fixed global classifier. After this training phase, the classifier head is fine-tuned for local adaptation. FedRep\cite{Collins2021FedRep} employs an alternating strategy: training the local classifier with a fixed feature extractor, followed by updating the feature extractor with the classifier held fixed. FedPAC\cite{Xu2023FedPAC} adopts a similar training approach. Additionally, it regularizes the feature space through feature prototypes, while also equipping each client with a personalized classifier combination.  While sharing some similarities, we employ local prototypes, with the aim of promoting personalization, not regularization. pFedFDA\cite{McLaughlin2024pFedFDA} assumes a latent distribution of client features. In each update round, pFedFDA estimates the local data distribution, derives a classifier head based on this estimate, and refines it via a local-global interpolation. Among these, our method is most similar to pFedFDA, under the shared assumption of a Gaussian representation space. However, pFedFDA still employs a fully-connected classifier head, which is estimated from local data. In contrast, our classifier head is formulated within a Gaussian Mixture framework and learned through model training. Unlike mainstream representation learning-based methods, our work focuses on estimating the local prototypes for each client to achieve personalized representation training, rather than personalized-classifier-head guidance.

\section{Preliminaries}
Consider a FL system comprising a central server and $M$ clients $C_i, \. i = 1,2, \dots,M$. Each client $C_i$ has model parameters $\bm{\theta}_i$ and a local training set $ \mathcal{D}_i=\{ (x_{i,j}, y_{i,j}) \}_{j=1}^{N_i} $. The FL objective can be formulated as:
\begin{equation} \label{eq_target}
	\min_{\bm{\theta}_1, \dots, \bm{\theta}_M \in \mathcal{Q}}  f(\bm{\theta}_1, \dots, \bm{\theta}_M )  \stackrel{\mathrm{def}}{=} \sum_{i=1}^M \alpha_i F_i (\bm{\theta}_i),
\end{equation}
where $\mathcal{Q}$ is the feasible set of model parameters, $F_i (\bm{\theta}_i)$ is the loss function associated with client $C_i$, $\alpha_i$ is the weight coefficient, and typically $\sum_{i=1}^M \alpha_i = 1$. In traditional FL, $\bm{\theta}_1 = \bm{\theta}_2 = \cdots = \bm{\theta}_M$, whereas in PFL, each client maintains personalized parameters. 

In PFL based on representation learning, most approaches employ a dedicated client classifier head, where this customized component guides the personalized training of the feature extractor. They formulate the following optimization objective via parameter decoupling:
\begin{equation} \label{eq_bilevel_opt}
	\min_{\bm{\phi} \in \Phi} \sum_{i=1}^M \alpha_i \min_{\bm{\psi}_i \in \Psi} F_i (\bm{\phi}, \, \bm{\psi}_i),
\end{equation}
where, $\bm{\psi}_i$ is the classifier head with distinct parameters for each client, and $\bm{\phi}$ is the shared feature extractor. Despite achieving good performance in most practices, configuring an excellent personalized classifier head remains challenging.


\begin{figure}[htbp]
	\centering
	\includegraphics[width=0.45\textwidth]{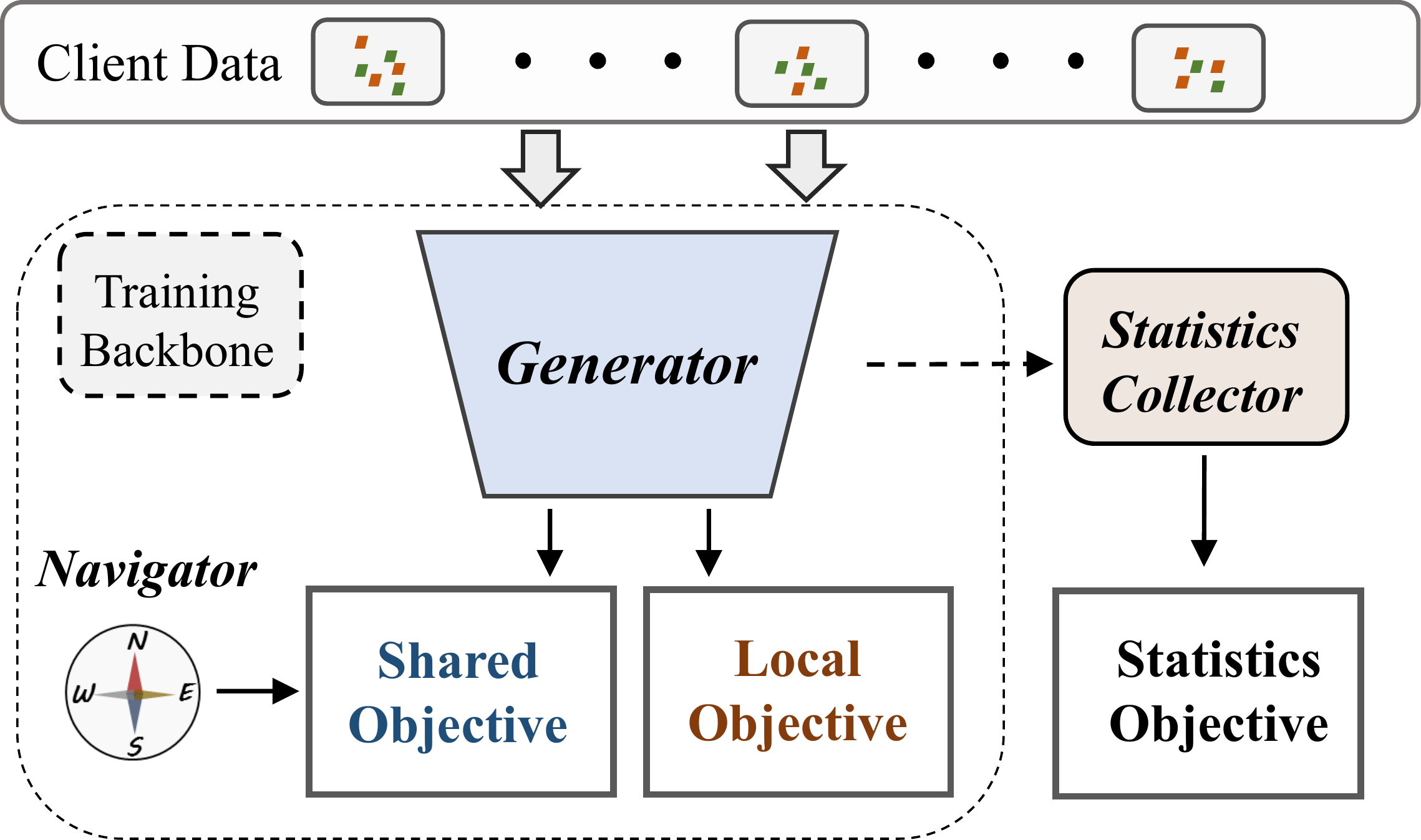}  
	\caption{ Global-local collaborative training module. In this process, the shared and local objectives jointly optimize the generator. The navigator defines and is refined by the shared objective, while a statistics extractor captures global statistics. }
	\label{fig_gengrator_framework}
\end{figure}

To mitigate this, we aim to first train a global generator (feature extractor), with its personalization guided by a personalized objective function. The shared objective is optimized based on the FedAvg algorithm, with local targets adapted via local statistical features. They are coupled within a Gaussian mixture framework, leading to the collaborative optimization of network parameters.

The global–local collaborative training module is illustrated in Fig. \ref{fig_gengrator_framework}. To promote a shared training objective for generators across different clients, we introduce a navigator that records the global navigation direction, thereby defining the overall direction of global optimization. Meanwhile, the local objective determines the client-specific optimization direction. These two directions are jointly leveraged to guide the update of generator parameters. In addition, a statistics extractor is employed to capture global statistical features, which facilitate the construction of subsequent classifiers.
﻿

The training objective of the backbone can be formulated as follows. Let client $c$ be a random variable taking on values $i$. In our proposed framework, $\bm{\phi}$ continues to denote the generator parameters, while $\bm{\psi}$ represents a well-designed navigator.  The event $\{c=i\}$ denotes that client $C_i$ is selected, whereby the federated learning objective can be written as:
\begin{equation} \label{main_loss}
	\min_{\bm{\phi} \in \Phi, \bm{\psi} \in \Psi} {\cal L} \stackrel{\mathrm{def}}{=}  \mathbb{E}_c \left[ \mathbb{E} \left[ \mathcal{H} (\bm{\phi}, \, \bm{\psi}) \,| \, c \right] + \lambda \cdot  \mathcal{R}_c (\bm{\phi})  \right],
\end{equation}
where $\mathcal{H}(\cdot)$ is the globally shared objective function, and $\mathcal{R}_c(\cdot)$ is the client-dependent personalized objective. This equation can be simplified to the following form:
\begin{equation}
	\min_{\bm{\phi} \in \Phi, \bm{\psi} \in \Psi}   \mathbb{E} \left[ \mathcal{H} (\bm{\phi}, \, \bm{\psi}) \right] + \lambda \cdot \mathbb{E}_c \left[   \mathcal{R}_c (\bm{\phi}) \right].
\end{equation}
In this expression, the first term promotes global collaborative training guided by the class labels, while the second term promotes local personalization based on the representation distribution. 

After training the generator, we customize the classifier head for each client based on the representation space distribution. It is implemented by fusing dual-scale information from global and local statistics. The specific details will be elaborated in the next section.

\section{Method}
\subsection{Gaussian Mixture and Re-sampling}
The key challenge in PFL lies in balancing collaborative training with personalization. However, in classification tasks, the information-rich raw signal is overly compressed into categorical labels. The resulting sparse representation is insufficient to capture both the global commonality and inter-client heterogeneity necessary for effective collaboration. To achieve effective classification while also modeling data heterogeneity, we train a generator to map the raw signal into a representation space. 

In this work, we place a Gaussian assumption on the global representation space for each class. In the case of a multi-class classification task, this naturally leads to a Gaussian mixture. Denote by $\bm{\mu}_k$ and $\bm{S}_k$ the mean and covariance matrix of the Gaussian distribution corresponding to the $k$-th class. Its probability density function is then given by $ f_k(\bm{z}) = \mathcal{N}(\bm{z}; \, \bm{\mu}_k, \bm{S}_k) $. For $K$ classes, the probability density function of the Gaussian mixture model can be expressed as:
\begin{equation}
	f(\bm{z}) = \sum_{k=1}^K \bm{\pi}_k \cdot \mathcal{N} (\bm{z}; \, \bm{\mu}_k, \bm{S}_k)
\end{equation}
where $\bm{\pi}_k$ is the weight coefficient for the $k$-th class.

Therefore, given the distribution of each class representation, the probability that a sampled $x$ belongs to the $k$-th class can be computed as:
\begin{equation} \label{eq_class}
	\mathcal{P}_k = \frac{\bm{\pi}_k f_k(\bm{z})}{f(\bm{z})} = \frac{ \bm{\pi}_k \cdot \mathcal{N}(\bm{z}; \, \bm{\mu}_k, \bm{S}_k)}{\sum_{i=1}^K \bm{\pi}_i \cdot \mathcal{N} (\bm{z}; \, \bm{\mu}_i, \bm{S}_i)},
\end{equation}
where $\bm{z} = \bm{z}(x; \, \bm{\phi}) $ is the value of $x$ in the representation space.

Then, we consider the representation for data of the same class but across different clients. The modeling of client data heterogeneity is achieved via simulated re-sampling. In other words, the distribution of each client is obtained by weighted re-sampling from the original distribution. Furthermore, we assume that the re-sampling weight is proportional to the Gaussian density. Then, the distribution  for each class of each client in the representation space remains Gaussian. Specifically, let $\mathcal{Z} \sim \mathcal{N} (\bm{z}; \,\bm{\mu}, \bm{S})$ be a multivariate Gaussian random variable. Consider $N$ i.i.d. realizations $\mathcal{Z}_1, \mathcal{Z}_2, \dots, \mathcal{Z}_N$. Then, sampling a point $\hat{Z}$ from this set with probability proportional to the probability density function of another multivariate Gaussian distribution $\mathcal{N} (\bm{z}; \, \bm{\nu}, \bm{\Omega})$, i.e., the probability of selecting $\mathcal{Z}_i$ is:
\begin{equation}
		\mathcal{P}(\hat{\mathcal{Z}} = \mathcal{Z}_i | \mathcal{Z}_{1:N}) =  \frac{\mathcal{N} (\mathcal{Z}_i; \, \bm{\nu}, \bm{\Omega})}{ \sum_{j=1}^N \mathcal{N} (\mathcal{Z}_j; \, \bm{\nu}, \bm{\Omega}) }.
	\end{equation}
Then $\hat{\mathcal{Z}}$ is also a multivariate random variable, and in the limit as $N \to \infty$, $\hat{\mathcal{Z}}$ is distributed as a multivariate Gaussian:
\begin{equation}
		\hat{Z} \sim \mathcal{N} (\bm{z}; \, \bm{\mu}^*, \bm{S}^*),
	\end{equation}
where $\bm{S}^* = (\bm{S}^{-1} + \bm{\Omega}^{-1})^{-1}, \bm{\mu}^* = \bm{S}^* ( \bm{S}^{-1} \bm{\mu} + \bm{\Omega}^{-1} \bm{\nu} )$. This result follows from the fact that the distribution of the weighted re-sampled points asymptotically converges to a distribution which is proportional to the product of two Gaussian densities. This resulting distribution is itself Gaussian, and its parameters are given by the standard Bayesian update rules \cite{Bishop2006pattern}. Our re-sampling scheme is an instance of the weighted bootstrap \cite{Barbe2012weighted}.

Therefore, for multiple classes, the representation space for each client remains a Gaussian mixture. Compared to the global distribution, each client's Gaussian components exhibit distinct means and covariance, with varying mixture weight coefficients. We attribute these differences to the representational heterogeneity arising from client data heterogeneity.

\subsection{Shared and Local Objectives}
Next, with the global and client representation distributions established, we propose the objective for PFL. The goal of supervised deep learning classification is to first ensure separability on training data and then generalize to general test instances. In our generative modeling framework, the separability of training data can be reduced to the separability of their representations. Under the Gaussian mixture distribution assumption, there are two pathways to make the representations of training data easily separable. The first approach is to push the class means as far apart as possible, while the second is to keep the variance as small as possible. Both contribute to sharper decision boundaries between classes. As shown in Fig. \ref{fig_pFedGM} (middle), we formulate the task of driving representations of different classes apart in distinct directions as a collaborative multi-client task optimized through the shared objective. Due to data heterogeneity across clients, their feature representations follow different distributions. To adapt to this, we frame the task of aggregating features of the same class toward a center as a client-specific personalized task, which is optimized via a personalized objective, as illustrated in Fig. \ref{fig_pFedGM} (right). 

During the training process, priority is given to learning the generator. First, consider the shared objective. Of course, a straightforward approach is to assign a global optimization direction to each class. Nevertheless, specifying a suitable direction and guiding the class representations accordingly is non-trivial. Inspired by the probabilistic calculation in Eq.~\eqref{eq_class}, we propose an adaptive direction along with an adaptive navigator.



Combine parameters $\bm{\pi}_k$ and $|\bm{S}_k|^{\frac{1}{2}}$ into $\bm{b}_k = \log \frac{\bm{\pi}_k}{|\bm{S}_k|^{\frac{1}{2}}}$, and set $\bm{A}_i = \bm{S}_i^{-1}$, from which the negative log posterior probability can be computed as: 
\begin{equation} 
	 H(\bm{\phi}, \bm{\psi}, \bm{\nu})  \stackrel{\mathrm{def}}{=} -\log \frac{\bm{\rho}_y}{ \sum_{i=1}^K \bm{\rho}_i },
\end{equation}
where
\begin{equation*} 
	\bm{\rho}_i = \exp\left( -\frac{1}{2} ( \bm{z} - \bm{\mu}_i )^T \bm{A}_i (\bm{z} - \bm{\mu}_i) + \bm{b}_i \right),
\end{equation*}
and $\bm{\psi} = \{\bm{\mu}_i, \bm{b}_i\}_{j=1}^K$, $\bm{\nu} = \{ \bm{A}_i \}_{i=1}^K $.

\begin{figure}[htbp]
	\centering
	\includegraphics[width=0.45\textwidth]{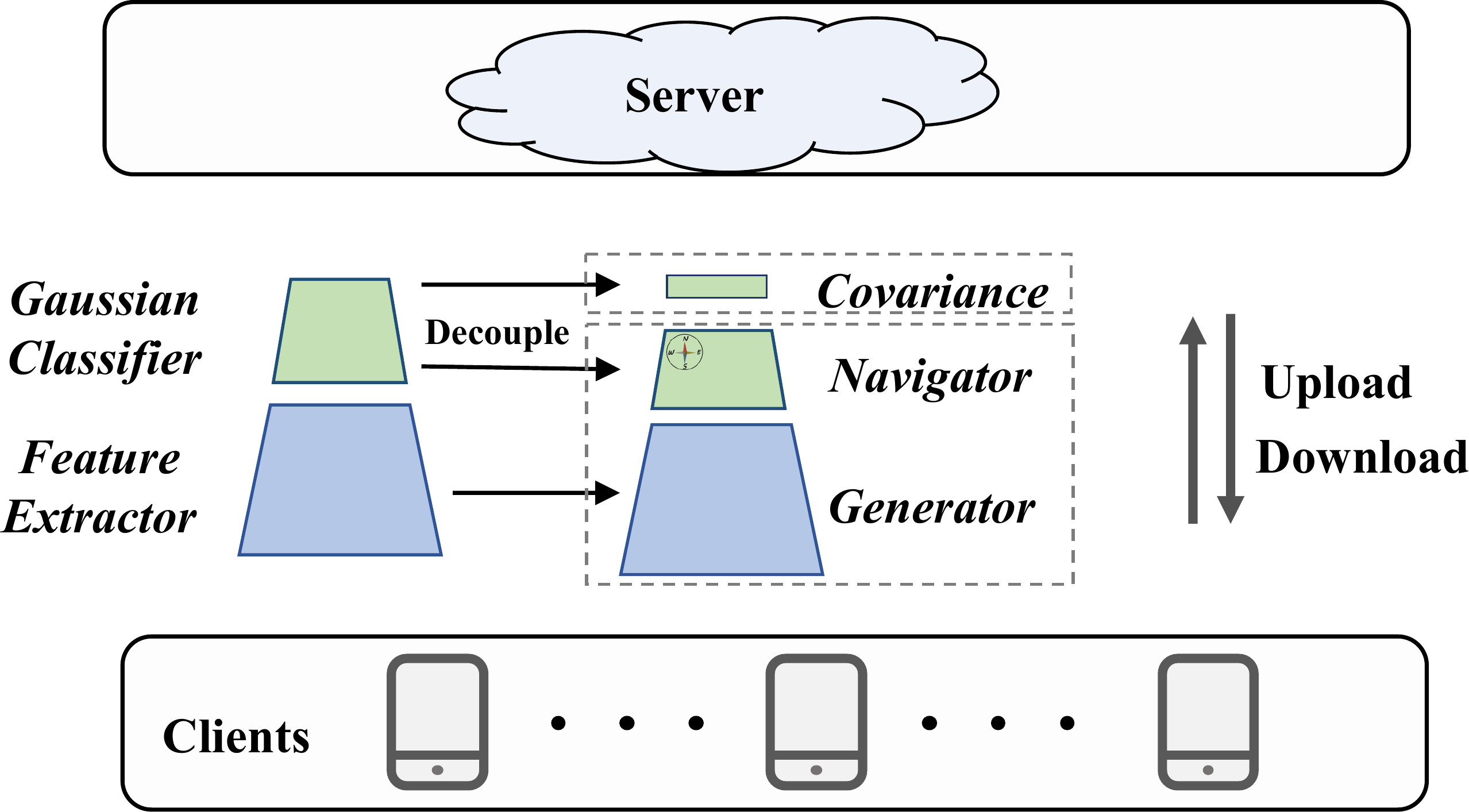}  
	\caption{Illustration of parameter decoupling. After decoupling the conventional Gaussian classifier into a navigator and covariance extractor (Covariance for short), these along with the generator are exchanged between server and clients. The navigator is utilized to generate the shared objective, and the covariance features are employed to craft the subsequent  personalized classifier heads. }
	\label{fig_decople}
\end{figure}

With $\bm{\phi}$ being trainable, minimizing the expected negative log posterior probability enables accurate classification for each training sample. However, due to covariance discrepancy, data either over-converges toward or diverges from class centroid, lacking the capability to properly drive inter-class feature divergence. Therefore, to fulfill the aforementioned shared objective, a decoupling strategy is applied to the parameters throughout model training. As shown in Fig. \ref{fig_decople}, the conventional Gaussian classifier is decoupled into a navigator and a covariance extractor. The navigator is constructed by fixing the covariance of the Gaussian classifier to the identity matrices. This results in the loss of covariance information. To compensate, a covariance extractor is introduced, the output of which is then used to configure the subsequent personalized classifier head.

Let $ \bar{\bm{E}} = \{\bm{E}\} $, and the shared objective is formulated accordingly:
\begin{equation} \label{eq_LossH}
	\mathcal{H}(\bm{\phi}, \bm{\psi})  \stackrel{\mathrm{def}}{=} H(\bm{\phi}, \bm{\psi}, \bar{\bm{E}}),
\end{equation}
where $\bm{\phi}$ denotes the generator parameters and $\bm{\psi}$ represents the navigator parameters. If we seek to optimize $\bm{\phi}$, the process involves backpropagating to feature representation $\bm{z}$ first, followed by calculating the derivative $\frac{\partial \bm{z}}{ \partial \bm{\phi}}$. As a result, minimizing this negative log posterior probability yields the partial derivative:
\begin{equation*}
	\frac{\partial \mathcal{H}}{ \partial \bm{z}} = -\bm{\mu}_y + \sum_{i=1}^K \mathcal{P}_i \bm{\mu}_i, 
\end{equation*}
where $ \mathcal{P}_i = \frac{\bm{\rho}_y}{ \sum_{i=1}^K \bm{\rho}_i } $ with $\bm{A}_i = \bm{E}$. 


Therefore, under gradient-based optimization, this effectively constrains the class representations to move along the direction of $\bm{\mu}_y - \sum_{i=1}^K \mathcal{P}_i \bm{\mu}_i$. Since $ \mathcal{P}_y < 1 $, the coefficient for $\bm{\mu}_y$ is positive while the others are negative. This makes the shared objective drive representations from different classes toward separate directions, consequently maximizing the distance between the deffierent class global training representations. We regard $\bm{\mu}_i$ as the navigational direction. The final moving direction of $\bm{z}$ is adaptively regulated by $\mathcal{P}_i$.


\begin{figure}[htbp]
	\centering
	\includegraphics[width=0.49\textwidth]{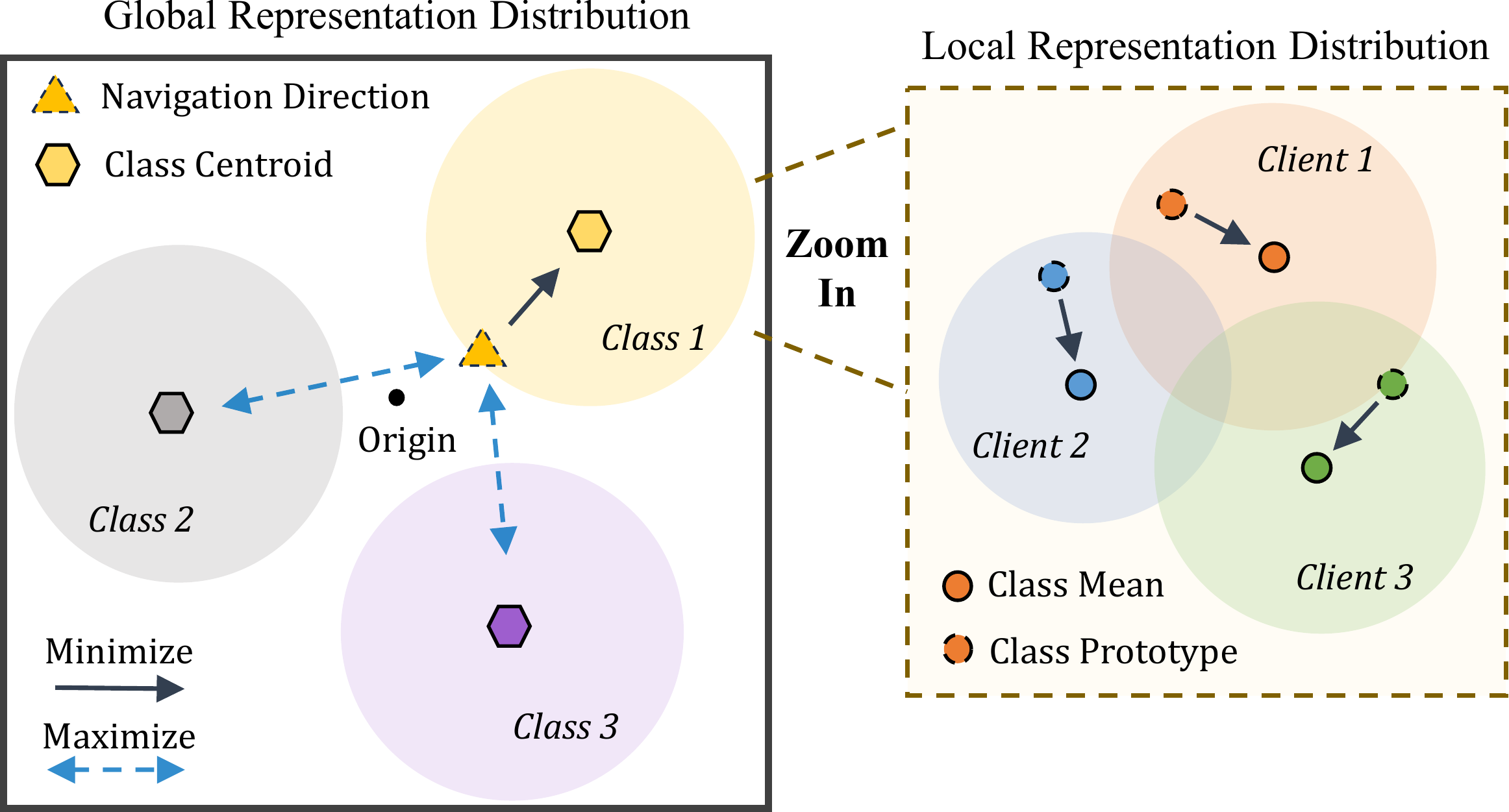}  
	\caption{ Navigational direction and local class prototype adaptation. (Left) The navigational direction ($\bm{\mu}_i$) self-adjusts based on the global representation distribution, with the objective of aligning with its own class centroid and diverging from others. (Right) The local class prototype ($\bm{\upsilon}_{i, k}$) continuously adapts toward the class mean. }
	\label{fig_Navigation}
\end{figure}

Next, we analyze the update of the navigational direction. With a considered $\bm{\mu}_i$ trainable parameter, the partial derivative of Eq.~\eqref{eq_LossH} with respect to $\bm{\mu}_i$ is given by:

\begin{equation*}
	\frac{\partial \mathcal{H}}{ \partial \bm{\mu}_y} = (1 - \mathcal{P}_y)(\bm{\mu}_y - \bm{z}), \,
	\frac{\partial \mathcal{H}}{ \partial \bm{\mu}_i} =  - \mathcal{P}_i(\bm{\mu}_i - \bm{z}), i \neq y.
\end{equation*}
Similar to the previous analysis, the navigation direction moves closer to samples of its own class while distancing itself from samples of other classes. The coefficient is also adjusted by $\mathcal{P}_i$. Fig. \ref{fig_Navigation} (left) illustrates the adaptive adjustment of the navigational direction. Here, we refer to the weighted sample center as the class centroid. The navigation direction for each class is adaptively guided by the optimization objective that pulls it toward its own class data and pushes it away from others. 


Eq.~\eqref{eq_LossH} does not account for the influence of covariance, implying that samples of all classes are assumed to have identical dispersion. However, due to differences in data distribution, the covariance of different classes typically differ after the generator is trained. To facilitate personalized adaptation, we optimize a global covariance after training the generator. To align with Eq.~\eqref{eq_LossH}, this can be achieved by minimizing the original negative log posterior estimate:
\begin{equation} \label{eq_cov}
	 \hat{\mathcal{H}}(\bm{\nu}) = H(\bm{\phi}^*, \bm{\psi}^*, \bm{\nu}) ,
\end{equation}
where $(\bm{\phi}^*, \bm{\psi}^*)$ are the optimized parameters. While post-training $\bm{\nu}$ after $\bm{\phi}$ and $\bm{\psi}$ are trained introduces extra communication rounds, the variable $\bm{\nu}$ is thus co-updated with $\bm{\phi}$ and $\bm{\psi}$ during optimization to approximately achieve the minimization. That is, each iteration minimizes:
\begin{equation} \label{eq_cov_approx}
	\hat{\mathcal{H}}(\bm{\nu}) \approx H(\bm{\phi}, \bm{\psi}, \bm{\nu}) ,
\end{equation}
where $\bm{\phi}$ and $\bm{\psi}$ are updated throughout the iterations by minimizing Eq.~\eqref{main_loss}. Assuming $H(\bm{\phi}, \bm{\psi}, \bm{\nu})$ is Lipschitz continuous with respect to $\bm{\phi}$ and $\bm{\psi}$, then as $\bm{\phi}$ and $\bm{\psi}$ approach convergence, we have $H(\bm{\phi}, \bm{\psi}, \bm{\nu}) \approx H(\bm{\phi}^*, \bm{\psi}^*, \bm{\nu}).$ Therefore, the update using Eq.~\eqref{eq_cov_approx} can be regarded as an approximation of Eq.~\eqref{eq_cov}. We further constrain $\bm{A}_i$ to be a diagonal matrix to reduce the number of parameters.



We now turn to the personalized objective. It focuses on reducing the variance  of the representation for each client and class (since the covariance matrix is constrained to be diagonal, the variances here refers to the variance of each individual element; similar usage applies henceforth). For a given client $c$ and class $y$, minimizing the variance can be computed as $\min \mathbb{E} \left[\lVert \bm{z} - \mathbb{E} [\bm{z} | y, c] \rVert^2 | y, c \right].$ Taking the expectation with respect to $y$ yields the personalized objective for each client:
\begin{equation} \label{eq_LossR}
	\mathcal{R}_c(\bm{\phi})  \stackrel{\mathrm{def}}{=}  \frac{1}{d} \mathbb{E} \left[  \lVert \bm{z} - \mathbb{E} [\bm{z} | y, c] \rVert^2  | c \right],
\end{equation}
where $d$ is the feature dimension used to scale the function value. Each client has a distinct aggregation objective. In this way, data features from different clients of the same class share a similar forward direction but exhibit distinct clustering patterns. Moreover, the client-wise clustering by class also aligns with maximizing Gaussian conditional probabilities, as concentrating features toward the class centroid increases the probability of that class.

Notably, $\mathbb{E} [\bm{z} | y, c]$ is defined as a function of $\bm{\phi}$. Consequently, minimizing Eq.~\eqref{eq_LossR} poses a challenge. However, the particular form of $\mathcal{R}_c(\bm{\phi})$ circumvents this requirement. Let $\bm{\upsilon} = \mathbb{E} \left[ \bm{z}|y,c \right]$, which is a random variable related to $y$ and $c$.  The following theorem describes this phenomenon.
\begin{theorem} \label{Thm_grad}
	Suppose at a certain step of a stochastic optimization algorithm, a client $c$ is selected. During client training, random sampling of local data yields samples $x_1, x_2, \dots, x_n$ and $\bm{z}_i = \bm{z}(x_i; \, \bm{\phi})$. Then,
	\begin{equation}
		\frac{\partial \mathcal{R}_c(\bm{\phi})}{\partial \bm{\phi}} =  \mathbb{E}  \left[  \frac{2}{nd} \sum_{i=1}^n   \frac{\partial \bm{z}_i}{\partial \bm{\phi}} \cdot \left( \bm{z}_i - \bm{\upsilon} \right) | c, \bm{\phi} \right].
	\end{equation}
\end{theorem}

This theorem reveals that when computing the gradient of $\mathcal{R}_c(\bm{\phi})$ with respect to $\bm{\phi}$, the $\bm{\upsilon}$ can be treated as a constant independent of $\bm{\phi}$. This property significantly reduces the computational cost. Nonetheless, computing $\bm{\upsilon} = \mathbb{E} \left[ \bm{z}|y,c \right]$ at every iteration remains somewhat complex. Instead, we calculate $\bm{\upsilon}$ precisely only at each communication round and adjust it using the samples from each iteration. Fig. \ref{fig_Navigation} illustrates the adjustment process of the $\bm{\upsilon}_{i,k}$ (class prototypes).


\subsection{Personalized Classifier Adaptation}
In this part, we operate under the assumption of a pre-trained global generator that projects each client's data onto a client-specific representation space. Meanwhile, the global navigator and the extracted covariance features are accessible. According to the assumptions, the representation space remains a Gaussian mixture. Therefore, the classification scheme shown in Eq.~\eqref{eq_class} remains applicable. While directly customizing a classifier head for each client could better fit their individual training data distributions, such highly personalized models would likely overfit due to the extremely limited local training samples, severely degrading generalization performance. A dual-scale fusion method, achieved through information gain, is employed to improve client-specific classification.

\subsubsection{Information Gain}

Inspired by the Kalman gain \cite{Simon2006Kalman}, we treat the global representation distribution as a prior estimate for the local one, and regard the local training data as the observations (likelihood). Then, according to Bayes' theorem, the posterior distribution is derived, which fuses the prior information with the observational data. Notably, because the framework is built upon Gaussian distributions, the resulting posterior remains Gaussian.

For a single class, we now have the global representation distribution $p(\bm{z})$, which also serves as the local prior. From the local training data, the locally observed distribution (likelihood) can be expressed as  $p(\bm{z}_c | \bm{z})$. Therefore, given the local prior and multiple i.i.d. observations $ \{\bm{z}_i\}_{i=1}^n$, the estimate of the state $\bm{z}$ can be inferred as:
\begin{align}
	p(\bm{z} | \bm{z}_1, \dots, \bm{z}_n) &\propto  p(\bm{z}) \cdot p(\bm{z}_1, \dots, \bm{z}_n | \bm{z})  \nonumber\\ 
	&= \underbrace{ p(\bm{z})}_{\text{prior}} \cdot \underbrace{p(\bm{z}_1 | \bm{z})\cdot p(\bm{z}_2 | \bm{z}) \cdots p(\bm{z}_n | \bm{z}) }_{\text{observed likelihood}}
\end{align}
Under the Gaussian assumption, the distribution of $p(\bm{z} | \bm{z}_1, \dots, \bm{z}_n)$ can be further expressed as:
\begin{align}
	\log p(\bm{z} | \bm{z}_1, \dots, \bm{z}_n) & =  -\frac{1}{2} ( \bm{z} - \bm{\mu}^* )^T \bm{A}^* (\bm{z} - \bm{\mu}^*) +  \nonumber\\ 
	& \sum_{i=1}^n -\frac{1}{2} ( \bm{z} - \bm{z}_i )^T \bm{A} (\bm{z} - \bm{z}_i) + \bm{b},
\end{align}
where $\bm{A}$ is the  inverse of the covariance matrix induced by observation noise, and $\bm{b}$ is a constant. This common covariance across observations allows the latter terms to be combined, leading to the following reformulated expression
\begin{align} \label{eq_single_gauss}
	\log p(\bm{z} | \bm{z}_1, \dots, \bm{z}_n) = & -\frac{1}{2} ( \bm{z} - \bm{\mu}^* )^T \bm{A}^* (\bm{z} - \bm{\mu}^*)  \nonumber\\ 
	&  -\frac{1}{2} ( \bm{z} - \bar{\bm{z}} )^T \bm{A}^{\prime} (\bm{z} - \bar{\bm{z}}) + \bm{b}^{\prime},
\end{align}
where $\bm{A}^{\prime} = n \cdot \bm{A}$, $\bm{b}^{\prime} = \bm{b} + \sum_{i=1}^n  (\bm{z}_i^TA\bm{z}_i -  \bar{\bm{z}}^TA\bar{\bm{z}})$ is a constant and $\bar{\bm{z}} = \frac{1}{n} \sum_{i=1}^n \bm{z}_i $. 

Given that the clients' observations are independent, the following holds for $K$ classes:
\begin{equation} \label{eq_K_gausss}
	\mathcal{P}_k = \frac{\bm{\pi}_k \cdot p_k(\bm{z} | \bm{z}_{k,1}, \dots, \bm{z}_{k, n_k}) }{ \sum_{i=1}^K \bm{\pi}_i \cdot p_i(\bm{z} | \bm{z}_{i,1}, \dots, \bm{z}_{i, n_i})},
\end{equation}
where $\mathcal{P}_k$ denotes the probability that $\bm{z}$ belongs to the $k$-th class. Under the Gaussian assumption, substituting Eq.~\eqref{eq_single_gauss} into Eq.~\eqref{eq_K_gausss} yields a global-local information fusion framework.



\subsubsection{Personalized Training}
\begin{figure}[htbp]
	\centering
	\includegraphics[width=0.49\textwidth]{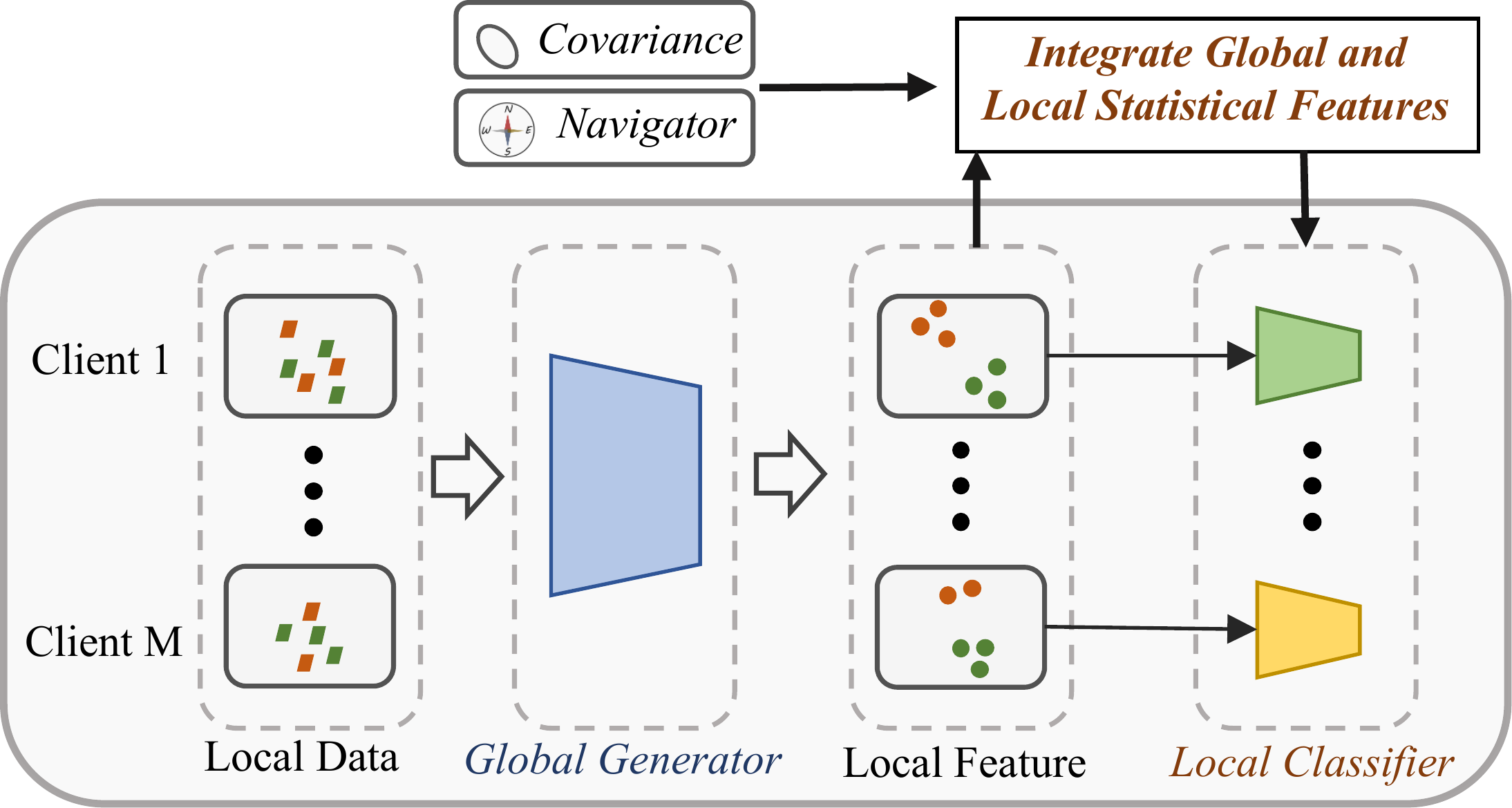}  
	\caption{Local classifier adaptation. Global information, encompassed by the navigator and covariance, is fused with local information for the joint construction of the local classifier. }
	\label{fig_model_framework}
\end{figure}

While the information gain approach was established in the previous subsection, key elements such as the determination of $\bm{A}^{\prime}$ and $\bm{b}^{\prime}$ remain unresolved. In this subsection, we determine the final local classifier through variable parameterization and adjustment. The framework for local classifier adaptation is shown in Fig. \ref{fig_model_framework}. The local data of each client is transformed into local feature representations via the global generator. Subsequently, each client acquires a local classifier based on its local features, the global covariance and the navigator (which contains global mean information and class weights).


For a given client $c$ and class $y$, let $\bm{\upsilon}_{i}^* = \mathbb{E} [\bm{z} | y = i, c]$. Following the previous derivation, the personalized fine-tuning loss is formulated by integrating the shared objective with its client-specific personalized objective:
\begin{equation}  \label{eq_LossC}
	{\cal L}_c(\bm{\psi}, \mathcal{A}_g, \mathcal{A}_c)  = \mathbb{E} \left[ -\log \frac{\exp\left( \bm{\xi}_y + \bm{\zeta}_y + \bm{\beta}_y \right)}{\sum_{i=1}^K \exp \left( \bm{\xi}_i + \bm{\zeta}_i + \bm{\beta}_i \right)} \right],
\end{equation}
where
\begin{equation*}
	\bm{\xi}_i = -\frac{1}{2} ( \bm{z} - \bm{\mu}_i^* - \bm{\mu}_i )^T \mathcal{A}_g \bm{A}_i^* (\bm{z} - \bm{\mu}_i^* - \bm{\mu}_i) ,
\end{equation*}
\begin{equation*}
	\bm{\zeta}_i =  -\frac{\lambda}{d} ( \bm{z} - \bm{\upsilon}_{i}^*  )^T \mathcal{A}_c (\bm{z} - \bm{\upsilon}_{i}^*) ,
\end{equation*}
\begin{equation*}
	\bm{\beta}_y =  \bm{b}_i^* +  \bm{b}_i ,
\end{equation*}
and $\bm{\psi} = \{ \bm{\mu}_i, \bm{b}_i \}_{i=1}^K$, $\bm{A}^*_i$,  and $\bm{b}^*_i$ are the global diagonal inverse covariance matrix and bias obtained after collaborative training, respectively. $\mathcal{A}_g$ and $\mathcal{A}_c$ are further constrained to be diagonal matrices. The generator parameters remain fixed as the global training result $\bm{\phi}^*$, while parameters $\bm{\psi}$, $\mathcal{A}_g$ and $\mathcal{A}_c$ are initialized as $0$, $\bm{E}$ and $ \bm{E}$, respectively, and fine-tuned for several epochs to adapt to the client representations. Notably, instead of directly fine-tuning the global and personalized covariance, we achieve this through appropriate reparameterization and imposed constraints. This is motivated by the need to simplify the classifier configuration and, more importantly, to mitigate the risk of overfitting.

The local personalized loss ${\cal L}_c(\bm{\psi}, \mathcal{A}_g, \mathcal{A}_c)$ functions as a trade-off between the global and local representation parameters. Global knowledge $\bm{\xi}_i$ and local knowledge $\bm{\zeta}_i$ are integrated via a coupling mechanism parameterized by $\bm{\psi}$, $\mathcal{A}_g$ and $\mathcal{A}_c$. The resulting representation remains consistent with a Gaussian mixture distribution, with the probability given by:
\begin{equation} \label{eq_ctest}
	\mathcal{P}_k =  \frac{\exp\left( \bm{\xi}_y + \bm{\zeta}_y + \bm{\beta}_y \right)}{\sum_{i=1}^K \exp \left( \bm{\xi}_i + \bm{\zeta}_i + \bm{\beta}_i \right)}.
\end{equation}
The updated covariance, mean and bias are given by:
\begin{equation*}
	\hat{\bm{S}}_i = \left( \mathcal{A}_g \bm{A}_i^* + \frac{2}{d} \mathcal{A}_c \right)^{-1},
\end{equation*}
\begin{equation*}
	\hat{\bm{\mu}}_i = \hat{\bm{S}}_i \cdot \left( \mathcal{A}_g \bm{A}_i^* (\bm{\mu}_i^* + \bm{\mu}_i) + \frac{2}{d} \mathcal{A}_c \bm{\upsilon}_i^* \right),
\end{equation*}
\begin{align*}
	\hat{\bm{b}}_i =& \, \bm{b}_i^* + \bm{b}_i + (\bm{\mu}_i^* + \bm{\mu}_i)^T \mathcal{A}_g \bm{A}_i^* (\bm{\mu}_i^* + \bm{\mu}_i) \\
	 &+ \frac{2}{d} (\bm{\upsilon}_i^*)^T \mathcal{A}_c \bm{\upsilon}_i^* - \hat{\bm{\mu}}_i^T \bm{S}^{-1} \hat{\bm{\mu}}_i.
\end{align*}
With labels, maximizing the class posterior probability yields the corresponding parameter estimates. 

Finally, we employ a granular adaptation strategy, adjusting $\bm{b}_i$ for each client $C_i$ to account for client-specific class bias. The parameter $\bm{b}_i$ originates from the weight coefficients of a Gaussian Mixture Model and is highly correlated with the class proportions of each client. Due to the high heterogeneity in class proportions across clients, parameter $\bm{b}_i$ exhibits significant variations. Fine-tuning $\bm{b}_i$ with all local data can mitigate the impact of such class imbalance. For each client, we use Eq.~\eqref{eq_LossC} as the loss function and perform fine-grained optimization using an off-the-shelf quasi-Newton method (e.g., L-BFGS\cite{Liu1989LBFGS}).

\subsection{Federated Optimization Algorithm}
With both the global and personalized objectives defined, we now consider the design of the federated optimization algorithm. The model training is a collaborative process across all clients, primarily aimed at learning a robust generator, while also obtaining auxiliary global parameters to prepare for subsequent personalized adaptation. Algorithm \ref{alg_pfedcm_detailed} describes the training workflow of pFedGM.

The proposed algorithm follows a two-phase optimization paradigm designed to balance global robustness with personalized adaptability. In Phase 1, all participating clients collaboratively train a shared global model under a Gaussian mixture prior, while estimating client-specific feature statistics for later personalization. The server maintains global parameters $\bm{\phi}$ (generator), $\bm{\psi}$ (navigator), and $\bm{\nu}$ (covariance), which are distributed to a subset of active clients each round. On each client, local updates are performed using a regularized objective that encourages inter-class separation and intra-class clustering, while simultaneously aggregating covariance information via the decoupled navigator–extractor structure. 

Phase 2 is executed independently on each client once global training concludes. Relying on the frozen generator from Phase 1, the client first extracts a compact representation of its local data. These representations then adapt the classifier via a mixture personalized objective to refine decision boundaries. A key step in this phase is the fine-grained adjustment of the bias terms $\{\bm{b}_1, \bm{b}_2, \cdots, \bm{b}_K\}$ using L-BFGS, enabling rapid client-specific calibration without overfitting. This strategy of fixing the generator while personalizing the classifier ensures the model retains globally learned representations while specializing efficiently to local data distributions.

﻿
Theorem \ref{Thm_grad} reveals that the partial derivative $\frac{\partial \mathcal{R}_c(\bm{\phi})}{\partial \bm{\phi}}$ can be computed without requiring $\frac{\partial \bm{\upsilon}(\bm{\phi})}{\partial \bm{\phi}} $, where $\bm{\upsilon}(\bm{\phi}) = \mathbb{E} \left[ \bm{z}|y,c \right]$. Accordingly, in the algorithm \ref{alg_pfedcm_detailed}, we estimate the value of $\bm{\upsilon}$ (i.e. $\bm{\upsilon}(\bm{\phi})$) to bypass its complex gradient computation. In fact, under the assumptions that the computed value of $\bm{\upsilon}$ is accurate and the variance of the personalized gradient estimate is bounded, the convergence of the local SGD iterative scheme can be established by following the proof technique and functional assumptions of \cite{Li2020Convergence}. The following theorem presents the convergence guarantees under full and partial client participation.

\begin{theorem} \label{Thm_app1}
	Under the assumptions that $\bm{\upsilon}$ is computed accurately, the personalized gradient estimate has bounded variance, and the objective functions $\{F_i\}_{i=1}^M$ along with the training samples satisfy certain regularity conditions, the local SGD iteration scheme converges as follows in the full or a specific partial client participation settings.
	\begin{equation}
		\mathbb{E}[f(\bar{\bm{\theta}}_T)] - f^* = O(\frac{1}{T}) ,
	\end{equation}
	where $ \bar{\bm{\theta}}_T = \sum_{i=1}^M \alpha_i \bm{\theta}_i^T $, $\bm{\theta}_i = \{\bm{\phi}_i, \bm{\psi}_i\}$, and $T$ denotes the number of iterations.
\end{theorem}



\begin{algorithm} 
	\caption{Workflow of pFedGM algorithm.}
	\label{alg_pfedcm_detailed}
	\vspace{5pt}

	\tcp{$\blacktriangleright$ \textbf{Phase 1: Global Collaborative Training}}  
	
	\vspace{5pt}

	Server initializes parameters $\bm{\phi}^0, \bm{\psi}^0$ with random Gaussian weights, and initializes $\bm{\nu}^0 = \bar{\bm{E}}$ \\
	
	\For{each round $r = 0$ \KwTo $R-1$}{
		
		Randomly select active clients and send $\bm{\phi}^r, \bm{\psi}^r, \bm{\nu}^r $ to them
		
		\For{each active client $i$}{
			$\bm{\phi}_{i}^{r} \leftarrow \bm{\phi}^{r}$,
			$\bm{\psi}_{i}^{r} \leftarrow \bm{\psi}^{r}$,
			$\bm{\nu}_{i}^{r} \leftarrow \bm{\nu}^{r}$ \\
			
			\For{each class $k$ in parallel}{ 
				Compute $ \bm{\upsilon}_{i, k}^r = \mathbb{E} \left[\bm{z}^r \, | y=k, c=i \right]$ 
			}
			\For{$j=0$ \KwTo local epochs}{ 
				\For{$s=0$ \KwTo end}{
					Sample $\mathcal{D}_i^{r,j,s} \sim \mathcal{D}_i$ \\
					Update $\bm{\nu}_{i}^{r}, \bm{\phi}_{i}^{r}, \bm{\psi}_{i}^{r}$ using loss \ref{eq_cov_approx} and loss \ref{main_loss} with $\mathbb{E} \left[\bm{z}^{r,j,s} \, | y=k, c=i \right]$ replaced by $\bm{\upsilon}_{i, k}^r $ \\
					Update $\bm{\upsilon}_{i}^r$ using $\mathcal{D}_i^{r,j,s}$ with a stepsize
				}
			}
			Send $ \bm{\phi}_{i}^{r},  \bm{\psi}_{i}^{r},  \bm{\nu}_{i}^{r}$ to the server
		}
		
		Server updates $\bm{\phi}_{g}^{r+1}, \bm{\psi}_{g}^{r+1}, \bm{\nu}_{g}^{r+1}$ using a weighted average of each client's $ \bm{\phi}_{i}^{r}$, $ \bm{\psi}_{i}^{r}$, $ \bm{\nu}_{i}^{r} $
	} 
	\Return{$\bm{\phi}^* = \bm{\phi}_{g}^R, \bm{\psi}^* = \bm{\psi}_{g}^R, \bm{\nu}^* = \bm{\nu}_{g}^R$}
	
	\vspace{12pt}
	\tcp{$\blacktriangleright$ \textbf{Phase 2: Personalized Client Adaptation}} 
	\vspace{5pt}
	Client initializes parameters $ \bm{\psi} = 0, \mathcal{A}_g = \bm{E}, \mathcal{A}_c = \bm{E}$
	
	\For{$x$ in local data}{
		Comput $\bm{z} = \bm{\phi}^*(x)$
	}
	\For{each class $k$ in parallel}{ 
		Compute $ \bm{\upsilon}_{k}^* = \mathbb{E} \left[\bm{z} \, | y=k \right]$
	}
	
	\For{$j=0$ \KwTo personalized epochs}{ 
		\For{$s=0$ \KwTo end}{
			Sample $\mathcal{D}^{j,s} \sim \{\bm{z}\}$ \\
			Update $\bm{\psi}, \mathcal{A}_g, \mathcal{A}_c$ using loss \ref{eq_LossC} \\
		}
	}
	Update $\{ \bm{b}_1, \bm{b}_2, \cdots, \bm{b}_K\} \in {\bm{\psi}}$ using L-BFGS with loss \ref{eq_LossC} for several steps
	
	\Return{$\hat{\bm{\psi}}^* = \bm{\psi}, \mathcal{A}_g^* = \mathcal{A}_g, \mathcal{A}_c^* = \mathcal{A}_c$}
	
\end{algorithm}

\begin{figure*}[htbp]
	\centering
	\includegraphics[width=1\textwidth]{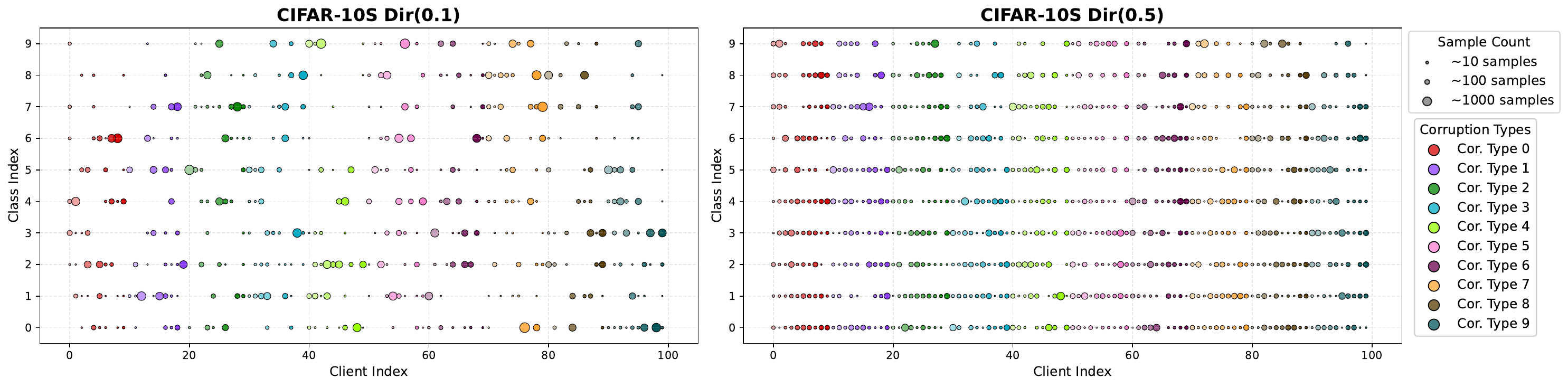}  
	\caption{The client data distribution of CIFAR-10S (all data corrupted). Different shades of the same color represent different levels of corruption, with darker shades indicating more severe corruption. }
	\label{fig_Dir}
\end{figure*}

\section{Experiments}
\subsection{Experimental Setup}

We compare pFedGM with the following baseline methods: Local training, where each client trains its model entirely on its own local data; FedAvg\cite{McMahan2017FedAvg} and FedAvgFT, which fine-tunes locally based on the global FedAvg model; multiple PFL methods, including FedPer\cite{Arivazhagan2019FedPer}, pFedMe\cite{Dinh2020pFedMe}, LG-FedAvg\cite{Liang2020LGFedAvg}, Ditto\cite{Li2021Ditto}, FedRep\cite{Collins2021FedRep}, FedBABU\cite{Oh2022FedBABU}, FedPAC\cite{Xu2023FedPAC}, and pFedFDA\cite{McLaughlin2024pFedFDA}. 

\subsubsection{Data Partitioning}

To validate the effectiveness of our method on image classification tasks, we conduct experimental evaluations on five popular datasets: EMNIST (handwritten characters, 28×28 grayscale); CIFAR-10/100 (natural images, 32×32 color); and the more challenging TinyImageNet (natural images, 64×64 color). The EMNIST dataset (62 classes), an extension of the original MNIST, includes both handwritten digits and letters, comprising 814,255 grayscale images. For recognizing natural objects, the CIFAR-10 and CIFAR-100 datasets provide 60,000 color images across 10 and 100 classes, respectively. To evaluate models on a more challenging and scalable task, we employ the TinyImageNet dataset, which contains 120,000 color images spanning 200 classes. These datasets collectively provide a hierarchy of progressively increasing complexity for benchmarking image classification models.

First is the standard federated learning setup with heterogeneous data. To simulate prior probability shift and quantity skew on the EMNIST, CIFAR-10/100 and TinyImageNet, we partition the data according to a Dirichlet distribution ($\alpha \in \{0.1, 0.5\}$), following \cite{Xu2023FedPAC, McLaughlin2024pFedFDA}. A lower $\alpha$ corresponds to a higher degree of heterogeneity. Specifically, non-IID data for each client is generated by sampling from a Dirichlet distribution per class as in \cite{Lin2020FedDF, Yurochkin2019Bayesian}. The Dirichlet concentration parameter $\alpha$ governs the heterogeneity level: smaller $\alpha$ values produce more skewed per-client class distributions. We experiment with $\alpha=0.1$ (highly non-IID) and $\alpha=0.5$ (moderately non-IID) to evaluate our method under varying heterogeneity conditions. For each client’s data, $80\%$ is randomly selected as the training set, and the remaining $20\%$ serves as the test set.

To simulate the cross-device challenges encountered in real-world natural environments, we consider common sources of input noise for natural images. Similar to \cite{McLaughlin2024pFedFDA}, we corrupt the training and testing data of clients in CIFAR-10/100 by applying image corruptions, each with five severity levels as defined in \cite{Hendrycks2019corruption}. The noise sources include factors such as measuring devices and environmental conditions, which lead to ten distinct types of corruption, spanning blur (motion, defocus), noise (Gaussian, shot, impulse), weather (frost, fog), digital (JPEG), and illumination (brightness, contrast) effects. To simulate data heterogeneity in real-world settings, we generate CIFAR-10S/100S by corrupting the CIFAR-10/100. With five severity levels set for each corruption type, the corruption-severity pairs constitute 50 unique categories and we evaluate our method under two scenarios. In the first scenario, to simulate the case of partial client corruption, each of the first 50 clients is subjected to one unique corruption category, while the remaining 50 clients remain uncorrupted. In the second scenario, we simulate universal client corruption by assigning a unique corruption category to each pair of clients. 

Fig. \ref{fig_Dir} illustrates the per-client data distribution of CIFAR-10S under comprehensive corruption. Client indices have been reordered to cluster those sharing the same corruption type. The resulting visualization reveals a distribution structured to highlight the impact of varying corruption types and levels. The left subfigure (Dirichlet concentration parameter $\alpha=0.1$) exhibits high heterogeneity, with most clients dominated by samples from only a few classes of a single corruption category, reflecting strong non-IID characteristics. In contrast, the right subfigure ($\alpha=0.5$) shows a more balanced and mixed allocation of corrupted data across clients, indicating a comparatively more uniform data partition.



\subsubsection{Model Setup}

For the EMNIST and CIFAR-10/100 datasets, we employ 4-layer and 5-layer convolutional neural networks (CNNs) respectively, following the architectures used in \cite{Xu2023FedPAC, McLaughlin2024pFedFDA}.For TinyImageNet, we enhance the 5-layer CNN (used for CIFAR-10/100) by incorporating two dropout layers. And the feature dimension is scaled up accordingly to fit the image size and class number. In our algorithm, the final layer of the model differs from other methods, featuring the structure shown in fig. \ref{fig_gengrator_framework}. Notably, this backbone structure has the same number of parameters as a standard fully connected layer. However, the covariance matrices prepared for personalization, along with the personalization process itself, introduce additional trainable parameters. Nevertheless, by reducing the covariance to a diagonal form, the resulting computational overhead becomes negligible.

In the TinyImageNet CNN, dropout layers are inserted before each of the last two fully connected layers. For methods such as FedPAC that process features from the first fully connected layer, we evaluate performance using features both before and after the dropout layer, reporting the superior result.  In our method, the two added dropout layers are placed before the first fully connected layer and the navigator. The features used in Eq.~\eqref{eq_LossR} are extracted before the dropout layer. During the subsequent personalization phase, after global training concludes, these dropout layers are set to evaluation mode.


\subsubsection{Training Setup}
Unless otherwise specified, all methods are trained for 200 global communication rounds with 5 local epochs across all datasets. Optimization is performed using mini-batch SGD (learning rate=0.01, momentum=0.5, weight decay=5e-4). For the EMNIST dataset, the batch size is set to 16 and the number of clients is set to 1000, with a participation rate $q = 0.03$. For the other datasets, the batch size is set to 50 and the number of clients is set to 100, with a participation rate $q = 0.3$. 


Hyperparameters across methods are tuned within a specified range to determine appropriate settings. For pFedMe, we tune the parameter $\lambda$ over $[0.5, 1.0, 5.0, 10.0, 15.0]$ and set $\lambda=5.0$. The parameter $\mu$ in Ditto is selected from $[0.1, 0.5, 1.0, 2.0, 5.0]$ and fixed at $ \mu = 1.0 $. For FedPAC, we tune the parameter $\lambda$ over $[0.1, 0.5, 1.0, 2.0, 5.0]$ and set $\lambda=1.0$. For our method, the hyperparameter $\lambda$ is set to $\lambda=1.0$.

The steps for collaborative training and personalized adaptation in pFedGM are outlined in Algorithm \ref{alg_pfedcm_detailed}. For the personalization step, the number of personalized epochs is set to 5, and the SGD optimizer (momentum=0.5, weight decay=5e-4) is employed with a relatively high learning rate of 0.05. The L-BFGS optimizer is run with a learning rate of 0.05, and a maximum of 10 internal iterations, for a total of 5 full training cycles. The implementation will be made publicly available.



\begin{table*}[htbp]
	\centering
	\caption{Average (standard deviation) test accuracy (\%) on multiple datasets.}
	\label{tab:standard}
	\normalsize
	\begin{tabular}{l c c c c c c c c c}
		\toprule
		Dataset &  \multicolumn{2}{c}{EMNIST} & \multicolumn{2}{c}{CIFAR-10} & \multicolumn{2}{c}{CIFAR-100} & \multicolumn{2}{c}{TinyImageNet} \\
		\cmidrule(lr){2-3} \cmidrule(lr){4-5} \cmidrule(lr){6-7} \cmidrule(lr){8-9}
		Partition & Dir(0.1) & Dir(0.5) & Dir(0.1) & Dir(0.5) & Dir(0.1) & Dir(0.5) & Dir(0.1) & Dir(0.5) \\
		\midrule
		\rowcolor[gray]{0.9}
		Local       & 86.56 \rgba{12}   & 71.00 \rgba{11}   & 86.43 \rgba{13} & 58.64 \rgba{13} & 36.61 \rgba{8.9} & 15.13 \rgba{4.8} & 25.48 \rgba{7.0} & 7.07 \rgba{2.7} \\
		\rowcolor[gray]{0.9}
		FedAvg      & 82.51 \rgba{13}   & 83.61 \rgba{8.2}  & 53.13 \rgba{11} & 61.53 \rgba{6.7}  & 23.79 \rgba{5.4} & 24.59 \rgba{4.3} & 21.37 \rgba{4.2} & 19.54 \rgba{3.0} \\
		\midrule
		FedAvgFT    & 95.02 \rgba{5.3}  & 88.73 \rgba{5.6}  & 89.46 \rgba{10} & 74.13 \rgba{8.5}  & 49.47 \rgba{8.5} & 30.13 \rgba{4.5} & 41.15 \rgba{6.1} & 21.41 \rgba{3.1} \\
		Ditto       & 94.71 \rgba{5.9}  & 88.53 \rgba{5.7}  & 89.60 \rgba{10} & 73.60 \rgba{8.2}  & 49.54 \rgba{8.6} & 30.12 \rgba{4.9} & 42.30 \rgba{5.9} & 23.29 \rgba{3.4} \\
		FedBABU     & 91.43 \rgba{8.4}  & 82.85 \rgba{8.3}  & 87.63 \rgba{14} & 72.50 \rgba{9.1}  & 43.37 \rgba{8.4} & 26.13 \rgba{5.2} & 39.40 \rgba{5.6} & 20.77 \rgba{3.0} \\
		FedPAC      & 94.26 \rgba{6.1}  & 89.70 \rgba{5.9}  & 89.59 \rgba{11} & \underline{76.29 \rgba{8.3}}  & \underline{53.14 \rgba{7.0}} & 36.76 \rgba{5.5} & \underline{46.12 \rgba{5.6}} & \underline{27.69 \rgba{3.6}}\\
		FedRep      & 91.30 \rgba{7.6}  & 80.99 \rgba{7.9}  & 87.61 \rgba{12} & 65.62 \rgba{10}  & 38.81 \rgba{8.5} & 17.55 \rgba{4.2} & 26.05 \rgba{6.2} & 11.69 \rgba{2.4} \\
		LG-FedAvg   & 93.53 \rgba{6.3}  & 83.96 \rgba{6.9}  & 88.35 \rgba{11} & 64.39 \rgba{10}  & 37.84 \rgba{8.7} & 16.93 \rgba{4.4} & 30.31 \rgba{6.2} & 13.36 \rgba{2.4} \\
		FedPer      & 93.39 \rgba{6.5}  & 83.93 \rgba{7.0}  & 89.43 \rgba{9.9}  & 65.65 \rgba{9.9}  & 39.19 \rgba{8.7} & 16.79 \rgba{4.7} & 30.20 \rgba{6.1} & 13.46 \rgba{2.4} \\
		pFedMe      & 93.75 \rgba{6.0}  & 87.83 \rgba{5.6}  & 87.76 \rgba{11} & 70.54 \rgba{8.8}  & 43.74 \rgba{8.4} & 24.73 \rgba{4.7} & 33.78 \rgba{5.4} & 16.60 \rgba{2.8} \\
		pFedFDA     & \underline{95.55 \rgba{4.8}}  & \underline{90.06 \rgba{5.1}}  & \underline{90.01 \rgba{9.6}}  & 76.05 \rgba{7.6}  & 50.32 \rgba{7.5} & \underline{39.28 \rgba{4.9}} & 42.98 \rgba{6.1} & 24.17 \rgba{3.8} \\
		\midrule
		pFedGM     & \textbf{95.59 \rgba{4.9}} & \textbf{90.44 \rgba{5.0}} & \textbf{91.00 \rgba{9.4}} & \textbf{77.49 \rgba{7.1}} & \textbf{57.56 \rgba{7.3}} &\textbf{ 42.25 \rgba{5.8}} & \textbf{51.17 \rgba{5.6}} & \textbf{35.45 \rgba{3.7}} \\
		\bottomrule
	\end{tabular}
\end{table*}

\begin{table*}[htbp]
	\centering
	\caption{Average (standard deviation) test accuracy (\%) on CIFAR-10S/100S (environmental heterogeneity setting).}
	\label{tab:corruption}
	\normalsize
	\begin{tabular}{l c c c c c c c c}
		\toprule
		Dataset & \multicolumn{2}{c}{CIFAR-10S Dir(0.1)} & \multicolumn{2}{c}{CIFAR-10S Dir(0.5)} & \multicolumn{2}{c}{CIFAR-100S Dir(0.1)} & \multicolumn{2}{c}{CIFAR-100S Dir(0.5)} \\
		\cmidrule(lr){2-3} \cmidrule(lr){4-5} \cmidrule(lr){6-7} \cmidrule(lr){8-9}
		Corruption  & 50C. & 100C. & 50C. & 100C. & 50C. & 100C. & 50C. & 100C. \\
		\midrule
		\rowcolor[gray]{0.9}
		Local     & 85.80 \rgba{14} & 86.52 \rgba{13} & 57.82 \rgba{13} & 56.83 \rgba{13} & 36.04 \rgba{9.8} & 35.51 \rgba{10} & 14.13 \rgba{4.7} & 14.07 \rgba{4.7} \\
		\rowcolor[gray]{0.9}
		FedAvg    & 51.91 \rgba{13} & 50.34 \rgba{14} & 57.86 \rgba{10} & 55.63 \rgba{10} & 21.39 \rgba{6.9} & 19.85 \rgba{6.5} & 23.43 \rgba{5.3} & 20.99 \rgba{6.0} \\
		\midrule
		FedAvgFT  & 89.67 \rgba{10} & \underline{89.63 \rgba{9.5}} & 72.38 \rgba{8.2} & 71.47 \rgba{9.6} & 48.51 \rgba{8.3} & 46.99 \rgba{7.9} & 29.42 \rgba{5.4} & 27.90 \rgba{5.5} \\
		Ditto     & 88.67 \rgba{11} & 89.15 \rgba{9.9} & 72.01 \rgba{8.5} & 70.59 \rgba{9.2} & 48.11 \rgba{8.5} & 47.90 \rgba{8.6} & 30.22 \rgba{5.6} & 28.73 \rgba{5.9} \\
		FedBABU   & 87.00 \rgba{14} & 86.70 \rgba{13} & 69.67 \rgba{12} & 68.17 \rgba{10} & 41.05 \rgba{11} & 35.14 \rgba{10} & 24.21 \rgba{5.3} & 23.19 \rgba{5.5} \\
		FedRep    & 86.91 \rgba{12} & 86.86 \rgba{12} & 63.47 \rgba{11} & 63.13 \rgba{11} & 37.97 \rgba{9.0} & 37.40 \rgba{9.4} & 15.36 \rgba{4.1} & 14.92 \rgba{4.4} \\
		FedPAC    & \underline{89.76 \rgba{10}} & 89.48 \rgba{10} & \underline{74.70 \rgba{8.5}} & \underline{73.03 \rgba{8.8}} & \underline{51.63 \rgba{8.0}} & \underline{50.79 \rgba{8.7}} & 34.93 \rgba{6.2} & 32.01 \rgba{6.1} \\
		LG-FedAvg & 88.59 \rgba{11} & 88.15 \rgba{11} & 64.43 \rgba{10} & 64.30 \rgba{11} & 38.06 \rgba{9.1} & 37.08 \rgba{8.9} & 16.55 \rgba{4.5} & 16.23 \rgba{5.1} \\
		FedPer    & 88.60 \rgba{11} & 88.51 \rgba{10} & 64.51 \rgba{10} & 63.81 \rgba{11} & 39.79 \rgba{8.7} & 39.76 \rgba{8.8} & 16.91 \rgba{4.9} & 16.97 \rgba{4.8} \\
		pFedMe    & 87.52 \rgba{12} & 86.93 \rgba{13} & 67.59 \rgba{9.2} & 66.06 \rgba{9.8} & 43.67 \rgba{8.8} & 43.01 \rgba{8.6} & 24.49 \rgba{4.8} & 24.05 \rgba{5.2} \\
		pFedFDA   & 89.38 \rgba{9.8} & 89.44 \rgba{10} & 74.07 \rgba{9.0} & 71.98 \rgba{8.9} & 49.18 \rgba{7.9} & 46.81 \rgba{8.9} & \underline{37.79 \rgba{6.7}} & \underline{33.69 \rgba{7.5}} \\
		\midrule
		pFedGM    & \textbf{90.49 \rgba{9.3}} & \textbf{89.67 \rgba{9.5}} & \textbf{76.33 \rgba{8.2}} & \textbf{74.26 \rgba{8.4}} & \textbf{56.25 \rgba{8.3}} & \textbf{53.72 \rgba{8.3}} & \textbf{39.05 \rgba{6.5}} & \textbf{36.58 \rgba{6.6}} \\
		\bottomrule
	\end{tabular}
\end{table*}

\begin{figure*}[htbp]
	\centering
	\includegraphics[width=0.95\textwidth]{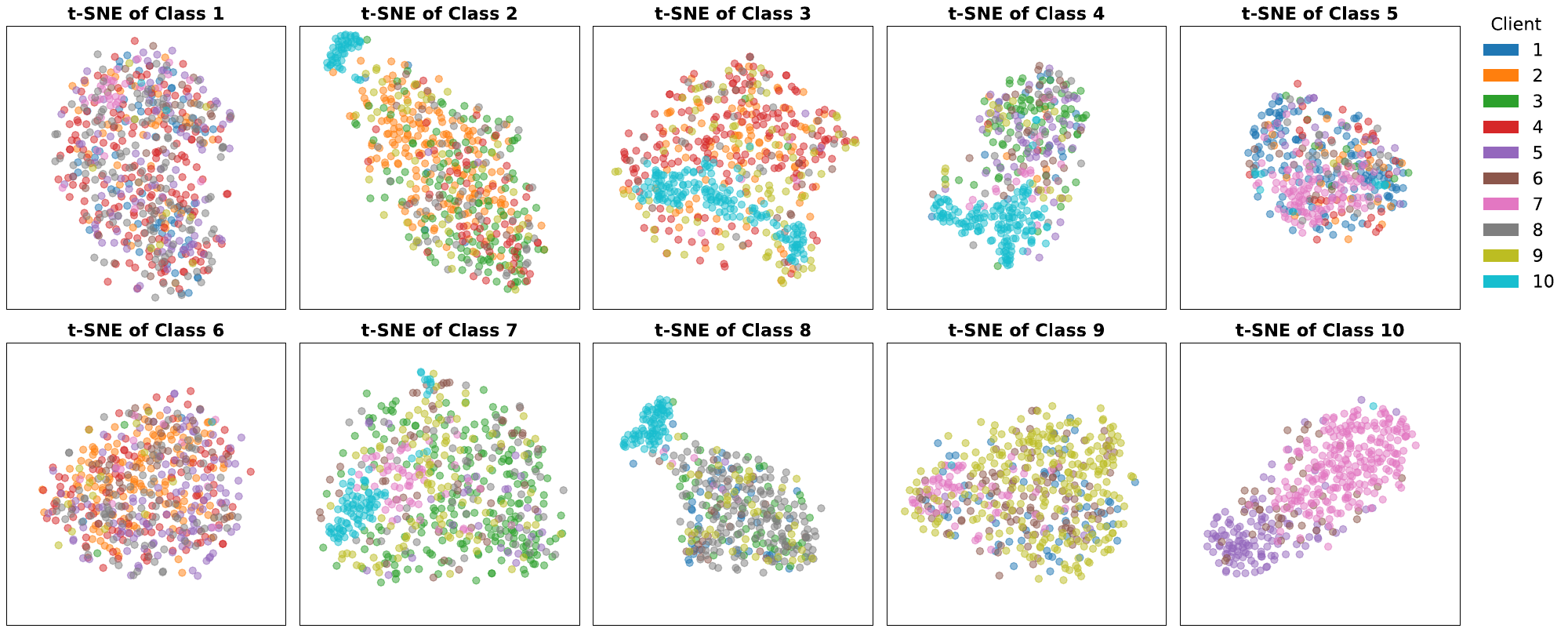}  
	\caption{t-SNE visualization of the feature representations. Each subplot depicts data from different clients of a single class in the representation space, revealing that clients within the same class exhibit distinct clustering structures.  }
	\label{fig_tsne}
\end{figure*}

\subsection{Numerical Results}

\subsubsection{Results Under Standard Settings}
We first evaluate different methods under standard settings and present the results in Table \ref{tab:standard}. Under this setup, client data heterogeneity stems solely from class imbalance (i.e., unequal numbers of samples per class). Here, Local is an algorithm trained exclusively on local data, and FedAvg is a classical non-PFL algorithm; they serve as the baselines. Dir($\alpha$) denotes different Dirichlet concentration parameters $\alpha$. Each cell in the table presents two values: the average test accuracy across all clients and the standard deviation (shown in parentheses), with both values scaled by 100 for readability.

Our method achieves superior test accuracy compared to other methods on EMNIST, CIFAR-10/100, and TinyImageNet, with its advantage being especially significant on the more challenging TinyImageNet dataset. Specifically, pFedGM consistently achieves state-of-the-art performance across most scenarios. On EMNIST, pFedGM holds only a slight advantage. In contrast, a more pronounced lead is observed on CIFAR-10, where it attains the highest accuracy under both Dir($0.1$) and Dir($0.5$) partitions ($91.00\%$ and $77.49\%$, respectively), outperforming the second-best methods (pFedFDA and FedPAC) by clear margins. This leading trend extends to CIFAR-100, where pFedGM reaches accuracies of $57.56\%$ and $42.25\%$, demonstrating a substantial improvement. The most compelling results are observed on TinyImageNet: pFedGM achieves top accuracies of $51.17\%$ ($\alpha=0.1$) and $35.45\%$ ($\alpha=0.5$), surpassing the second-best method, FedPAC, by significant margins of $+5.05\%$ and $+7.76\%$, respectively. This underscores the method's strong capability in handling complex data with substantial heterogeneity. Overall, the comprehensive results validate the robustness and general superiority of pFedGM in standard federated learning benchmarks.

\subsubsection{Results Under Environmental Heterogeneity Settings}
The evaluation of different approaches under environmental heterogeneity is presented in Table \ref{tab:corruption}. Here, 50C. and 100C. denote 50 and 100 corrupted clients, respectively. Across all experiments, our method achieves the highest test accuracy, demonstrating excellent robustness. Specifically, pFedGM exhibits stable and leading performance across various heterogeneous environments partitioned via a Dirichlet distribution (Dir($\alpha$)). As shown in the table, under a high-heterogeneity setting ($\alpha=0.1$), pFedGM surpasses all compared methods on both the CIFAR-10S and CIFAR-100S datasets, achieving accuracies of $90.49\%/89.67\%$ and $56.25\%/53.72\%$, respectively. In the relatively homogeneous setting ($\alpha=0.5$), pFedGM’s advantage is particularly pronounced on the more complex CIFAR-100S task, where it attains significantly higher accuracy ($39.05\%$ and $36.58\%$) than other PFL methods. This highlights its superior capability in handling complex, heterogeneous data patterns.

Fig. \ref{fig_tsne} presents a t-SNE visualization of the feature representations in the representation space. This is based on the CIFAR-10S dataset under the Dir($0.5$) partition, with 100 corrupted clients. Due to the limited number of test samples per client, the visualization is instead based on the training sets of ten clients, with each subplot representing a distinct class. The visualization reveals that, across multiple classes, feature representations from the same class maintain an overall clustering structure while displaying diversity across different clients.  

\begin{table*}[htbp]
	\centering
	\caption{Evaluation of new-client generalization on CIFAR-10 Dir(0.5).}
	\label{tab:generalize}
	\small 
	\setlength{\tabcolsep}{4.5pt}
	\begin{tabular}{l c c c c c c c c c}
		\toprule
		\multicolumn{2}{c}{Dataset}  & FedAvg & FedAvgFT & FedBAU & pFedMe & LG-FedAvg & FedPAC & pFedFDA & pFedGM \\
		\midrule
		\rowcolor[gray]{0.9}
		\multicolumn{2}{c}{Original Clients}   & 57.64 \rgba{7.8}  & 72.38 \rgba{8.4}  & 69.18 \rgba{9.5}  & 68.25 \rgba{9.0}  & 63.25 \rgba{11}   & 73.70 \rgba{8.7}  & \underline{73.96 \rgba{8.6}}  & \textbf{74.80 \rgba{7.3}}\\
		\midrule
		\multirow{11}{*}{\makecell{New \\ Clients}} 
		& clean       & 59.18 \rgba{3.9}  & \underline{72.31 \rgba{9.2}}  & 61.87 \rgba{19}   & 67.25 \rgba{12}   & 63.77 \rgba{17}   & 72.15 \rgba{10}   & \textbf{72.94 \rgba{11}}   & \textbf{72.94 \rgba{13}} \\
		& motion      & 49.08 \rgba{9.9}  & 68.62 \rgba{9.7}  & 62.31 \rgba{10}   & 64.92 \rgba{8.1}  & 58.00 \rgba{8.3}  & \underline{70.31 \rgba{7.1}}  & 68.31 \rgba{6.4}  & \textbf{70.92 \rgba{5.7}}\\
		& defocus     & 59.80 \rgba{6.0}  & 74.62 \rgba{8.5}  & 69.34 \rgba{4.3}  & 69.34 \rgba{8.6}  & 64.74 \rgba{10}   & \underline{74.28 \rgba{9.0}}  & 72.74 \rgba{9.5}  & \textbf{75.30 \rgba{6.3}}\\
		& gaussian    & 57.44 \rgba{6.9}  & 69.12 \rgba{12}   & 68.80 \rgba{12}   & 68.16 \rgba{13}   & 62.88 \rgba{14}   & \underline{73.76 \rgba{12}}   & 72.00 \rgba{12}   & \textbf{74.72 \rgba{8.9}}\\
		& shot        & 57.48 \rgba{3.5}  & 67.44 \rgba{9.0}  & 50.66 \rgba{20}   & 66.94 \rgba{6.8}  & 59.97 \rgba{9.3}  & \textbf{72.43 \rgba{7.7}}  & 70.60 \rgba{5.4}  & \underline{72.26 \rgba{6.1}}\\
		& impulse     & 53.76 \rgba{6.9}  & 68.18 \rgba{11}   & 60.50 \rgba{14}   & 66.61 \rgba{9.3}  & 59.25 \rgba{9.1}  & \underline{70.85 \rgba{9.7}}  & \underline{70.85 \rgba{10}}   & \textbf{73.20 \rgba{9.9}}\\
		& frost       & 41.23 \rgba{10}   & \underline{61.05 \rgba{9.1}}  & 56.26 \rgba{8.3}  & 59.91 \rgba{11}   & 47.61 \rgba{7.3}  & 58.09 \rgba{12}   & 58.54 \rgba{8.6}  & \textbf{61.50 \rgba{11}}\\
		& fog         & 45.20 \rgba{14}   & 69.54 \rgba{7.6}  & 69.37 \rgba{5.8}  & 69.87 \rgba{8.0}  & 65.23 \rgba{10}   & 70.53 \rgba{9.6}  & \textbf{71.52 \rgba{8.4}}  & \underline{71.36 \rgba{8.2}}\\
		& jpeg        & 57.53 \rgba{7.2}  & 67.43 \rgba{7.6}  & 66.21 \rgba{8.9}  & 66.82 \rgba{7.2}  & 60.58 \rgba{7.9}  & \underline{71.23 \rgba{5.7}}  & 70.93 \rgba{8.2}  & \textbf{74.89 \rgba{8.1}}\\
		& brightness  & 47.82 \rgba{8.6}  & 76.64 \rgba{8.6}  & 74.45 \rgba{9.2}  & 72.90 \rgba{10}   & 66.82 \rgba{12}   & 75.55 \rgba{8.5}  & \underline{77.41} \rgba{7.4}  & \textbf{78.66 \rgba{5.8}}\\
		& contrast    & 32.76 \rgba{14}   & 66.21 \rgba{15}   & 63.48 \rgba{19}   & 67.41 \rgba{13}   & 59.22 \rgba{18}   & 65.36 \rgba{17}   & \underline{67.58 \rgba{17}}   & \textbf{69.62 \rgba{14}}\\
		\bottomrule
	\end{tabular}
\end{table*}

\begin{figure*}[htbp]
	\centering
	\includegraphics[width=1\textwidth]{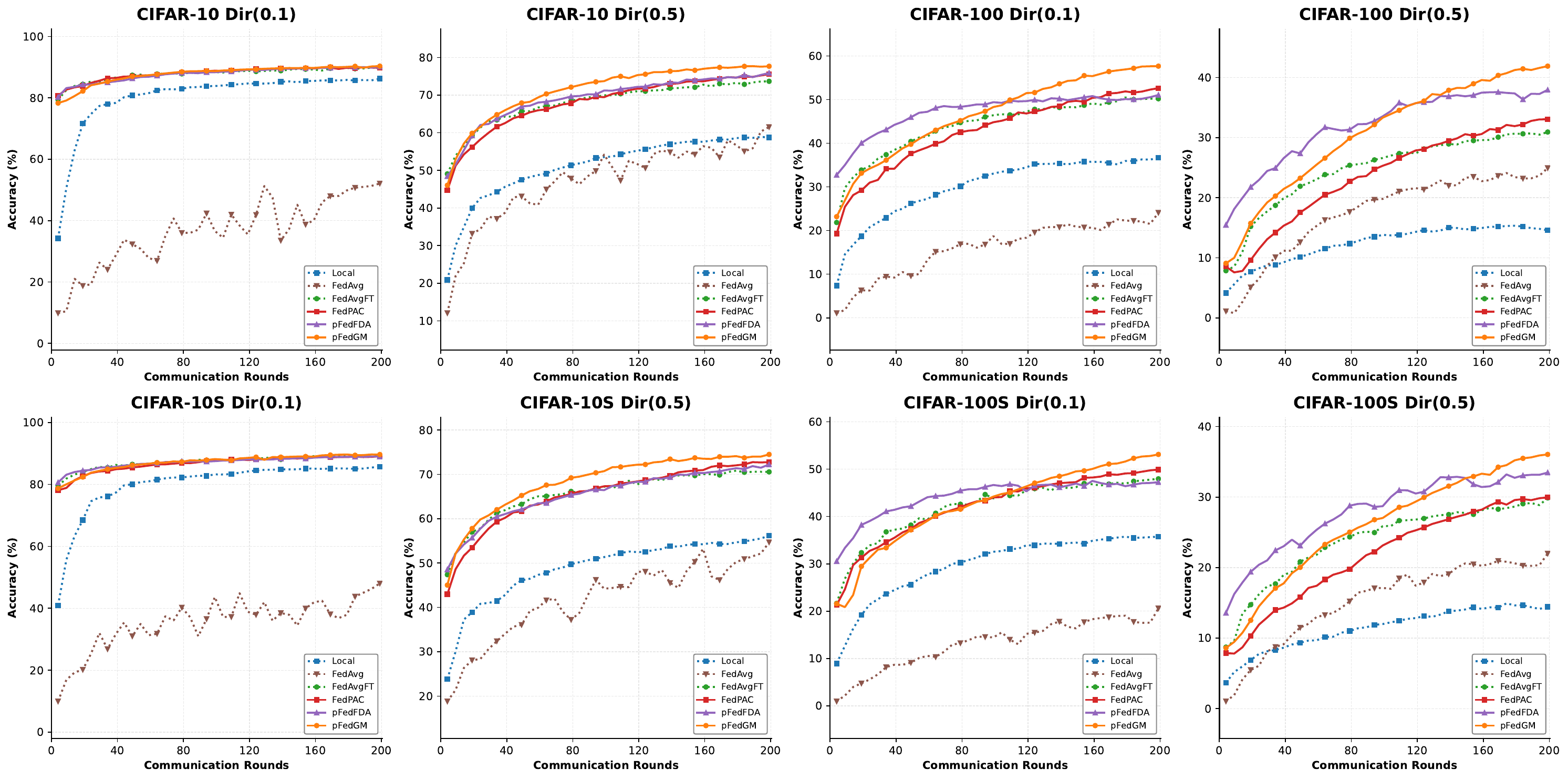}  
	\caption{The variation of test accuracy during the training process with sampling rate $q=0.1$. }
	\label{fig_analysis}
\end{figure*}

\begin{table*}[htbp]
	\centering
	\caption{Comparison of average (standard deviation) test accuracy (\%) under moderate settings.}
	\label{tab:client}
	\normalsize
	\begin{tabular}{l c c c c c c c c c}
		\toprule
		Dataset &  \multicolumn{2}{c}{EMNIST} & \multicolumn{2}{c}{CIFAR-10} & \multicolumn{2}{c}{CIFAR-100} & \multicolumn{2}{c}{TinyImageNet} \\
		\cmidrule(lr){2-3} \cmidrule(lr){4-5} \cmidrule(lr){6-7} \cmidrule(lr){8-9}
		Partition & Dir\rgba{0.1} & Dir\rgba{0.5} & Dir\rgba{0.1} & Dir\rgba{0.5} & Dir\rgba{0.1} & Dir\rgba{0.5} & Dir\rgba{0.1} & Dir\rgba{0.5} \\
		\midrule
		\rowcolor[gray]{0.9}
		Local      & 91.50\rgba{8.8} & 81.71\rgba{7.7} & 90.28\rgba{12} & 68.19\rgba{9.2} & 47.75\rgba{4.7} & 25.95\rgba{4.2} & 36.70\rgba{4.0} & 16.42\rgba{1.6} \\
		\rowcolor[gray]{0.9}
		FedAvg     & 83.18\rgba{14} & 85.25\rgba{7.9} & 55.60\rgba{13} & 68.25\rgba{5.8} & 31.28\rgba{6.0} & 35.15\rgba{2.8} & 27.91\rgba{4.3} & 29.56\rgba{1.7} \\
		\midrule
		FedAvgFT   & 96.16\rgba{4.7} & 90.79\rgba{4.8} & 91.87\rgba{8.3} & 79.14\rgba{6.3} & 58.46\rgba{3.9} & 43.44\rgba{2.9} & 50.13\rgba{3.1} & 36.57\rgba{1.6} \\
		FedPAC     & 96.03\rgba{4.5} & 90.74\rgba{4.8} & \underline{92.08\rgba{9.2}} & 79.32\rgba{6.2} & \underline{59.51\rgba{4.5}} & 44.87\rgba{3.4} & \underline{52.34\rgba{2.2}} & 36.21\rgba{1.8} \\
		pFedFDA    & \underline{96.43\rgba{4.3}} & \underline{91.45\rgba{4.6}} & 91.84\rgba{8.1} & \underline{79.65\rgba{6.5}} & 57.10\rgba{4.2} & \underline{47.60\rgba{2.8}} & 47.36\rgba{3.3} & \underline{37.15\rgba{1.6}} \\
		pFedGM      & \textbf{96.55\rgba{4.1}} & \textbf{91.82\rgba{4.5}} & \textbf{92.48\rgba{7.9}} & \textbf{80.93\rgba{5.5}} & \textbf{63.79\rgba{4.4}} & \textbf{51.04\rgba{2.9}} & \textbf{54.38\rgba{2.9}} & \textbf{42.35\rgba{1.7}} \\
		\bottomrule
	\end{tabular}
\end{table*}

\begin{table*}[htbp]
	\centering
	\caption{Comparison of system runtime (min) on the CIFAR-10/100 dataset.}
	\label{tab:Time}
	\normalsize 
	\setlength{\tabcolsep}{4.5pt}
	\begin{tabular}{l c c c c c c c c c c}
		\toprule
		Dataset & FedAvg & Ditto & FedPer & FedRep & FedBABU & pFedMe & LG-FedAvg & FedPAC & pFedFDA & pFedGM \\
		\midrule
		CIFAR-10  & 64.91  & 96.08  & 61.94  & 67.03  & 62.34  & 102.21 & 62.68  & 113.31 & 164.62 & 81.75  \\
		CIFAR-100 & 65.85  & 104.33 & 68.25  & 68.97  & 63.99  & 114.24 & 61.22  & 167.72 & 172.15 & 89.23  \\
		
		\bottomrule
	\end{tabular}
\end{table*}

\subsection{Additional Results}
\subsubsection{Generalization to New Clients}

To evaluate the adaptability of different methods to new clients, we split the clients into two groups and collaboratively train the model only on clients from one group. Using the CIFAR-10 Dir($0.5$) configuration, we generate 100 clients. Then we keep the first 50 clients with their clean data unchanged, while corrupting the remaining 50 clients with different types and severity levels of corruption, in the same manner as CIFAR-10S. The model is trained on the first 45 clients, with a participation rate $q = 0.6$. After training, the model is evaluated on 11 types of new clients: 10 corresponding to the corruption types, plus one clean type represented by the 5 remaining uncorrupted clients.

The evaluation results on different data types are shown in Table \ref{tab:generalize}. Our method demonstrates a clear lead on ``jpeg"-type corrupted data for new clients ($74.89\%$ vs. $71.23\%$, $+3.66\%$), and overall achieves the top generalization accuracy on 9 out of the 11 new evaluated data types,  while securing the second-highest accuracy on the remaining two. This demonstrates its overall advantage. The superior generalization performance underscores the robustness and adaptability of our method to previously unseen client data distributions, particularly under diverse and challenging corruption scenarios. This can be attributed to its generalizable representation learning, which prevents overfitting to the seen client distributions and promotes the learning of transferable features. Therefore, our approach not only excels in standard federated learning settings but also proves to be a more reliable solution for real-world applications where models must generalize to new clients with potentially heterogeneous and corrupted data.

\subsubsection{Results Under Limited Client Participation}
To investigate model stability under an extremely low client participation rate, we evaluate the performance of different methods with $q=0.1$. Experiments are conducted on CIFAR-10/100 and CIFAR-10S/100S, where CIFAR-10S/100S have all client data corrupted as described previously. The client participation rate for training is uniformly set to $q=0.1$, while all other settings remain unchanged. 

We illustrate the variation of test accuracy during the training process in Fig. \ref{fig_analysis}. As illustrated, under a low client participation rate ($q=0.1$), our proposed pFedGM consistently outperforms all compared methods across both the clean (CIFAR-10/100) and corrupted (CIFAR-10S/100S) datasets. This indicates that the generative modeling approach in pFedGM effectively captures the underlying data distribution while mitigating the adverse effects of corruption and statistical heterogeneity. The performance advantage is particularly pronounced under more challenging settings (e.g., CIFAR-100S), demonstrating the robustness of our method in federated environments with limited client participation and corrupted local data.

\subsubsection{Results Under Moderate Settings}
In this part, we provide additional test results for scenarios with a more moderate settings. The CIFAR-10/100, and TinyImageNet datasets are partitioned across 20 clients following a Dirichlet distribution ($\alpha \in \{0.1, 0.5\}$), while the EMNIST dataset is partitioned among 200 clients. Consequently, the data volume per client increases substantially. In response, we reduce the number of local training epochs to 2, with global rounds and other settings held constant. 

A comparison of test results across methods is provided in Table \ref{tab:client}. pFedGM demonstrates superior performance across all datasets and under both Non-IID partition schemes (Dir(0.1) and Dir(0.5)), achieving the best results in the vast majority of cases. Notably, its advantage over other methods becomes even more pronounced under the more challenging  datasets, CIFAR-100 and TinyImageNet. These results validate that pFedGM effectively mitigates client drift and overfitting, making it suitable for resource-constrained yet data-rich federated learning applications.

\subsubsection{Runtime Comparison}
Table \ref{tab:Time} presents the runtime under the standard configuration on CIFAR-10/100 with Dir($0.5$). During training, the additional time overhead of pFedGM compared to FedAvg primarily arises from computing the per-class means in each communication round and updating the global covariance parameter $\bm{\nu}$. Additionally, equipping each client with a personalized classifier head introduces a modest time overhead. However, this overhead occurs only once within the entire federated learning system and is not recurrent during the training of the backbone network. Consequently, the incurred additional time cost is negligible. Overall, despite introducing additional time overhead, our method maintains a considerable advantage over other well-performing methods such as FedPAC and pFedFDA.

\begin{table*}[htbp]
	\centering
	\caption{Ablation study on the classifier head. The ablation settings are denoted as follows: NA (No Adaptation) applies no personalized classifier head adaptation; FA (Fine-tuning Adaptation) employs only fine-tuning of the classifier head; and GA (Granular Adaptation) further incorporates a fine-grained adjustment of $\bm{b}_i$ following the fine-tuning step. }
	\label{tab:adaptation}
	\small 
	\begin{tabular}{l c c c c c c c c c c c c}
		\toprule
		\multicolumn{3}{c}{Ablation Strategy} &  \multicolumn{2}{c}{EMNIST} & \multicolumn{2}{c}{CIFAR-10} & \multicolumn{2}{c}{CIFAR-100} & \multicolumn{2}{c}{TinyImageNet} & \multirow{2}{*}{\makecell{Acc. Avg. \\ (Improv. Rate)}}   \\
		\cmidrule(lr){1-3} \cmidrule(lr){4-5} \cmidrule(lr){6-7} \cmidrule(lr){8-9} \cmidrule(lr){10-11}  
		NA & FA & GA & Dir(0.1) & Dir(0.5) & Dir(0.1) & Dir(0.5) & Dir(0.1) & Dir(0.5) & Dir(0.1) & Dir(0.5) & \\
		\cmidrule(lr){1-3} \cmidrule(lr){4-5} \cmidrule(lr){6-7} \cmidrule(lr){8-9} \cmidrule(lr){10-11} \cmidrule(lr){12-12}
		$\checkmark$ & &  & 80.86 & 83.58 & 55.32 & 63.59 & 25.32 & 30.52 & 23.52 & 26.31 & \cellcolor{blue!7}48.63 \rgbb{$0\% \uparrow$}  \\
		& $\checkmark$ &  & 95.08 & 90.12 & 90.88 & 77.02 & 53.40 & 41.44 & 48.46 & 34.72 & \cellcolor{blue!7} 66.39 \rgbb{$36.52\% \uparrow$} \\
		& & $\checkmark$  & \textbf{95.59} & \textbf{90.44} & \textbf{91.00} & \textbf{77.49} & \textbf{57.56} & \textbf{42.25} & \textbf{51.17} & \textbf{35.45} & \cellcolor{blue!7} \textbf{67.62} \rgbb{$1.85\% \uparrow$} \\
		
		\midrule
		
		\multicolumn{3}{c}{Ablation Strategy} & \multicolumn{2}{c}{CIFAR-10S Dir(0.1)} & \multicolumn{2}{c}{CIFAR-10S Dir(0.5)} & \multicolumn{2}{c}{CIFAR-100S Dir(0.1)} & \multicolumn{2}{c}{CIFAR-100S Dir(0.5)} & \multirow{2}{*}{\makecell{Acc. Avg. \\ (Improv. Rate)}} \\
		\cmidrule(lr){1-3} \cmidrule(lr){4-5} \cmidrule(lr){6-7} \cmidrule(lr){8-9}  \cmidrule(lr){10-11} 
		NA & FA & GA  & 50C. & 100C. & 50C. & 100C. & 50C. & 100C. & 50C. & 100C. & \\
		\cmidrule(lr){1-3} \cmidrule(lr){4-5} \cmidrule(lr){6-7} \cmidrule(lr){8-9} \cmidrule(lr){10-11} \cmidrule(lr){12-12}
		$\checkmark$ & &  & 49.87 & 47.13 & 60.70 & 57.33 & 23.33 & 19.71 & 25.58 & 22.38 & \cellcolor{blue!7} 38.25 \rgbb{$0\% \uparrow$}  \\
		& $\checkmark$ &  & 90.25 & 89.46 & 76.00 & 74.08 & 51.78 & 49.13 & 37.90 & 35.00 & \cellcolor{blue!7} 62.95 \rgbb{$64.58\% \uparrow$} \\
		& & $\checkmark$  & \textbf{90.49} & \textbf{89.67} & \textbf{76.33} & \textbf{74.26} & \textbf{56.25} & \textbf{53.72} & \textbf{39.05} & \textbf{36.58} & \cellcolor{blue!7} \textbf{64.54} \rgbb{$2.53\% \uparrow$}  \\
		
		\bottomrule
	\end{tabular}
\end{table*}

\begin{figure}[htbp]
	\centering
	\includegraphics[width=0.45\textwidth]{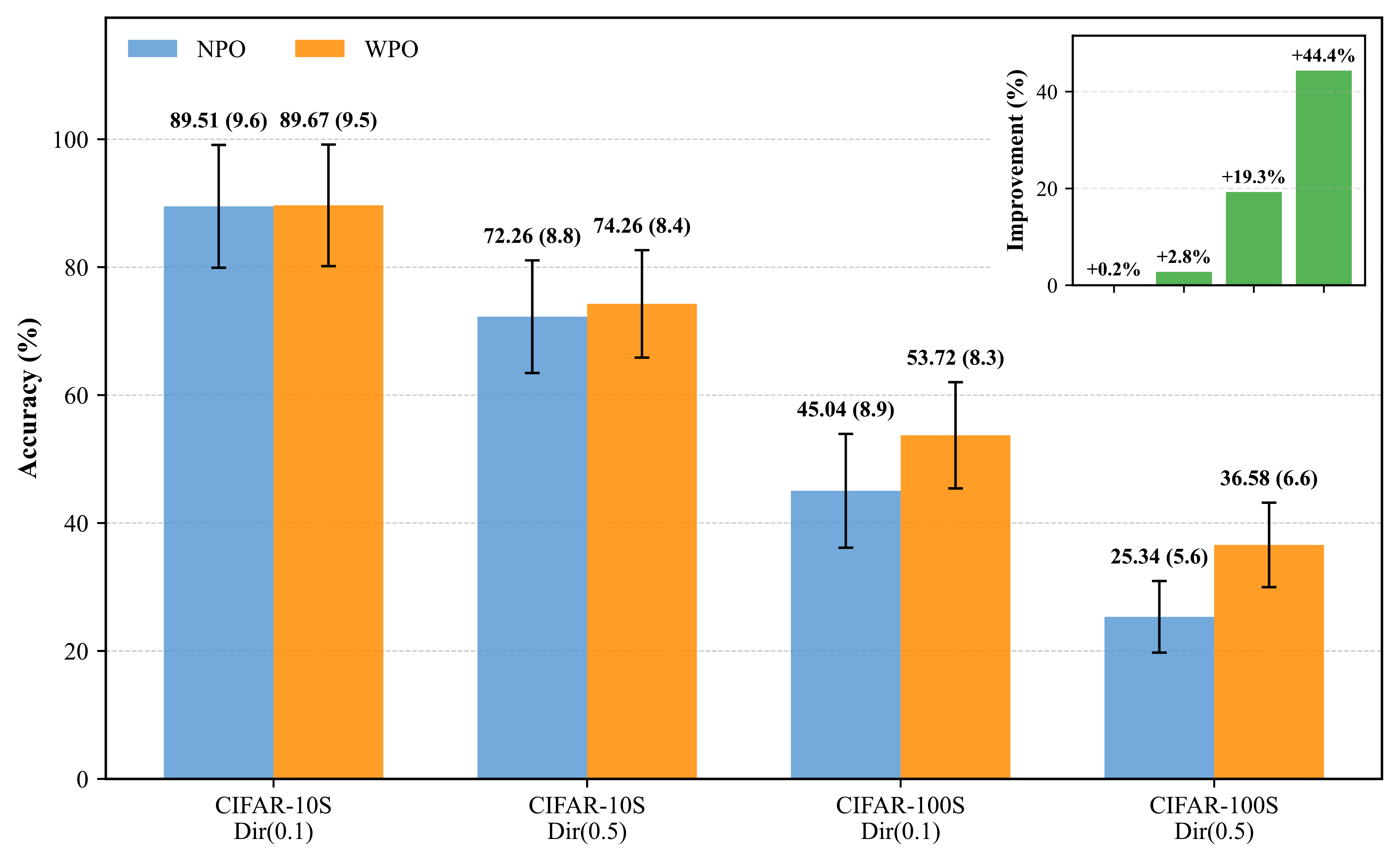}  
	\caption{Ablation study on the personalized objective function. NPO and WPO denote training without and with the personalized term, respectively.}
	\label{fig_weight}
\end{figure}

\begin{table}[htbp]
	\centering
	\caption{Ablation study on the influence of shared objective and statistic extractor. G.M. denotes training without the decoupling mechanism; G.M. - C. denotes using the decoupling mechanism but removing the statistic extractor; and G.M. + D. represents our proposed method.}
	\label{tab:decope}
	\small 
	\setlength{\tabcolsep}{2.5pt}
	\begin{tabular}{l c c c c}
		\toprule
		Dataset & \multicolumn{2}{c}{CIFAR-10S} & \multicolumn{2}{c}{CIFAR-100S}   \\
		\cmidrule(lr){2-5} 
		Partition & Dir(0.1) & Dir(0.5) & Dir(0.1) & Dir(0.5) \\
		\midrule
		G.M. & 89.32 \rgba{9.7} & 73.92 \rgba{8.5} & 52.46 \rgba{8.4} & 36.15 \rgba{6.4} \\ \addlinespace[0.2em]
		G.M. - C. & 89.63 \rgba{9.8} & 74.02 \rgba{8.3} & 51.65 \rgba{9.0} & 36.01 \rgba{6.6} \\ \addlinespace[0.2em]
		G.M. + D. & \textbf{89.67 \rgba{9.5}} & \textbf{74.26 \rgba{8.4}} & \textbf{53.72 \rgba{8.3}} & \textbf{36.58 \rgba{6.6}} \\
		\bottomrule
	\end{tabular}
\end{table}


\subsection{Ablation Studies}
\subsubsection{Impact of Classifier Adaptation}
To validate the effectiveness of the proposed method components, we conduct ablation studies. We first evaluate the impact of client-side classifier head adaptation by comparing the results under No Adaptation (NA), Fine-tuning Adaptation (FA), and Granular Adaptation (GA), as shown in Table \ref{tab:adaptation}. 

The ablation study reveals a substantial performance gap between the baseline without adaptation (NA) and the fine-tuning strategy (FA), with FA achieving an average accuracy improvement of $36.52\%$ and $64.58\%$ on standard and corrupted datasets, respectively. This demonstrates the essential role of personalizing the classifier head in handling data heterogeneity. Furthermore, the proposed granular adaptation (GA) strategy, which further introduces a fine-grained adjustment for the bias term $\bm{b}_i$, consistently outperforms the fine-tuning strategy approach across all experimental configurations. It provides an additional accuracy gain of $1.85\%$ on average for standard datasets and $2.53\%$ for corrupted datasets. Notably, the advantage of granular adaptation is more pronounced on more complex tasks (e.g., CIFAR-100, TinyImageNet) and under highly non-IID data partitions (Dir(0.1)). These results confirm that granular, client-specific refinement in the classifier head is not only effective but necessary, ultimately enabling pFedGM to achieve state-of-the-art performance.


\subsubsection{Effect of Personalized Objective}
The personalized objective not only customizes the aggregation center for each client but also actively contracts intra-class variance, thereby providing a clearer classification boundary for the subsequent Gaussian mixture-based framework. In this part, we examine the impact of incorporating the personalized objective function into the training process

The results are shown in Fig. \ref{fig_weight}. The two training variants are defined as: NPO (No Personalized Objective), where the model is trained without the personalized term $\mathcal{R}_c$; and WPO (With Personalized Objective), where the model is trained with the term $\mathcal{R}_c$. The results demonstrate that the inclusion of the personalized objective function yields consistent and substantial performance gains. Specifically, the relative accuracy improvements (highlighted in the inset) are most pronounced under the more challenging conditions (CIFAR-100S). On CIFAR-100S with a mild non-IID partition (Dir(0.5)), WPO achieves a significant $+44.4\%$ increase in accuracy compared to NPO. Even under the more imbalanced Dir(0.1) setting, the improvement remains notable at $+19.3\%$. This trend is consistent, though slightly less dramatic, on CIFAR-10S. The personalized objective, by contracting intra-class variance and customizing the aggregation center, effectively provides a clearer and more separable feature representation. This is critical for the downstream Gaussian mixture modeling, as it leads to more distinct and reliable component assignments, which in turn results in the superior generalization performance observed for pFedGM. The ablation confirms that the personalized objective is not merely beneficial but is a core component responsible for the method's robustness, especially in complex data environments.


\subsubsection{Influence of Shared Objective and Statistic Extractor}
In collaborative training, we decouple the Gaussian classifier into a navigator and a statistic extractor to achieve inter-class separation and covariance collection. This ablation study investigates the impact of this design choice, with results shown in Table \ref{tab:decope}. In the table, G.M. denotes training without the decoupling mechanism; G.M. - C. denotes using the decoupling mechanism but removing the statistic extractor; and G.M. + D. represents our proposed method, which employs the decoupling mechanism along with the statistic extractor. All other components remain unchanged, including the constraint that the covariance matrix is restricted to a diagonal matrix.

The results in Table \ref{tab:decope} demonstrate that the proposed G.M. + D. method consistently outperforms the other two variants across all settings, confirming the effectiveness of the full decoupling design. Its advantage is particularly pronounced on the more challenging CIFAR-100S dataset: it delivers an absolute improvement of $+1.26\%$ over the vanilla G.M. and $+2.07\%$ over G.M. - C. under the highly non‑IID Dir(0.1) partition. Notably, removing the statistic extractor (G.M. - C.) yields only negligible or even negative gains compared to the baseline. This shows that decoupling the classifier alone, without the covariance‑aware refinement, fails to capture the full benefit of the structural separation. Collectively, these findings validate that the synergistic integration of navigator‑guided separation and statistic‑informed refinement is crucial for achieving robust generalization in heterogeneous federated corruption scenarios.

\section{Conclusion and Future Work}
In classification-based federated learning, over-compressed information prevents data heterogeneity from being fully manifested in the output. Our work establishes that initiating federated personalization with a generative modeling phase, by first training a shared generator, provides a foundation for learning representations that are both discriminative and adaptable to client heterogeneity. By jointly refining client representation distributions through collaboration and personalization, the model benefits from global data while remaining well-adapted to local data. Finally, by integrating both global and local distributional features, each client is ultimately equipped with a personalized classifier capable of accurate, client-specific prediction. Its efficacy is supported by comprehensive experiments conducted in standard heterogeneous as well as environmentally heterogeneous scenarios. Future work includes exploring the scalability of pFedGM in more complex settings, and investigating other optimization objectives based on representation distributions.

\bibliographystyle{IEEEtran}
\bibliography{references}

\end{document}